\documentclass[10pt,journal,compsoc]{IEEEtran}
\newcommand{\scalarfordc}{1}
\usepackage{amsmath}
\DeclareMathOperator{\Tr}{Tr}
\DeclareMathOperator{\vectorize}{vec}

\usepackage{cuted}
\usepackage[draft]{hyperref}
\usepackage{array}
\usepackage{textcomp}
\usepackage{nicefrac}
\usepackage{amssymb}
\usepackage{amsfonts}
\usepackage{multirow}
\usepackage{textcomp}
\usepackage{epsfig}
\usepackage{epstopdf}
\newcommand{\datatensor}{\underline{\mathbf{X}}}
\newcommand{\datamat}{{\mathbf{X}}}
\newcommand{\collectedvar}{\bm{\Phi}}
\newcommand{\collecteddualvar}{\bm{\Psi}}
\newcommand{\errortensor}{\underline{\mathbf{E}}}
\newcommand{\anomaltensor}{\underline{\mathbf{O}}}
\newcommand{\anomalgraphmat}{{\mathbf{O}}}

\newcommand{\lagragian}{L}
\newcommand{\comunitycover}{\mathcal{C}}
\newcommand{\estcomunitycover}{\hat{\mathcal{C}}}
\newcommand{\comind}{c}
\newcommand{\covscalar}{\alpha}
\newcommand{\conductance}[1]{\pi{(#1)}}
\newcommand{\coverage}[1]{\chi{(#1)}}

\newcommand{\cover}{\mathcal{S}}
\newcommand{\estcover}{\hat{\mathcal{S}}}
\newcommand{\estassignment}{\hat{\assignment}}
\newcommand{\estassignmentvec}{\hat{\assignmentvec}}
\newcommand{\factmatforcom}{\mathbf{C}}
\newcommand{\entropy}[1]{\text{H}(#1)}
\newcommand{\mutualinfo}[2]{\text{I}(#1,#2)}
\newcommand{\datatensormiss}{\datatensor^M}
\newcommand{\datatensormissest}{\hat{\datatensor}^M}
\newcommand{\datatensoravail}{\datatensor^A}

\newcommand{\factmat}{\mathbf{A}}
\newcommand{\factnot}[1]{_{#1}}
\newcommand{\factind}{n}
\newcommand{\eye}{\mathbf{I}}

\newcommand{\dualadmmmat}{\mathbf{Y}}
\newcommand{\dualadmmvec}{\mathbf{y}}

\newcommand{\diagmat}{\mathbf{D}}
\newcommand{\parafacnot}[3]{\big[[{#1},{#2},{#3}]\big]}
\newcommand{\graphmat}{\mathbf{G}}
\newcommand{\graphvec}{\mathbf{g}}
\newcommand{\graphmatmiss}{\graphmat^M}
\newcommand{\graphmatmissest}{\hat{\graphmat}^M}
\newcommand{\graphmatavail}{\graphmat^A}
\newcommand{\commnbr}{C}
\newcommand{\assignment}{\alpha}
\newcommand{\assignmentvec}{\bm{\assignment}}

\newcommand{\elementwiseproduct}{\ast}
\newcommand{\regpar}{\mu}

\newcommand{\ranktensor}{R}
\newcommand{\ranktensorind}{r}
\newcommand{\regadmm}{\rho}
\newcommand{\katriraofac}{\mathbf{M}}
\newcommand{\factmatest}{\hat{\factmat}}
\newcommand{\cfactmatest}{\hat{\cfactmat}}
\newcommand{\pfactmatest}{\hat{\pfactmat}}

\newcommand{\diagvecest}{\hat{\diagvec}}
\newcommand{\pdiagvecest}{\hat{\pdiagvec}}
\newcommand{\ordertensor}{N}
\newcommand{\tensind}{i}
\newcommand{\factorsize}{I}
\newcommand{\diagvec}{\mathbf{d}}
\newcommand{\pdiagvec}{\tilde{\diagvec}}
\newcommand{\cfactmat}{\bar{\factmat}}
\newcommand{\pfactmat}{\tilde{\factmat}}

\newcommand{\posconstfun}[1]{g(#1)}
\newcommand{\transpose}{\top}

\newcommand{\errorgraphmat}{\mathbf{V}}
\newcommand{\misstensoperator}{\mathcal{P}_{\underline{\Omega}}}

\newcommand{\admmiter}{l}
\newcommand{\admmiternot}[1]{^{#1}}

\newcommand{\missmatoperator}[1]{\mathcal{P}_{\Omega_{#1}}}
\newcommand{\fitfun}{f}

\usepackage{fixltx2e}

\usepackage{graphicx}

\usepackage{subcaption}              

\usepackage[utf8]{inputenc}

\usepackage{amsfonts}
\usepackage{amsmath}
\usepackage{mathtools}

\usepackage{amssymb}

\usepackage{bm}

\usepackage{color,verbatim}
\usepackage{multirow}
\usepackage{accents}

\usepackage{theoremref}

\usepackage{flushend}

\usepackage{url}

\newcounter{rulecounter}
\newcommand{\resetrule}{ \setcounter{rulecounter}{0}}
\resetrule

\newsavebox{\selvestebox}
\newenvironment{colbox}[1]
  {\newcommand\colboxcolor{#1}%
   \begin{lrbox}{\selvestebox}%
   \begin{minipage}{\dimexpr\columnwidth-2\fboxsep\relax}}
  {\end{minipage}\end{lrbox}%
   \begin{center}
   \colorbox{\colboxcolor}{\usebox{\selvestebox}}
   \end{center}}

\definecolor{orange}{rgb}{1,0.8,0}
\definecolor{gray}{rgb}{.9,0.9,0.9}
\definecolor{darkgray}{rgb}{.3,0.3,0.3}
\definecolor{darkblue}{rgb}{.1,0.0,0.3}
\definecolor{lightblue}{rgb}{0.7,0.7,1}
\definecolor{lightred}{rgb}{1,0.7,.7}
\definecolor{purple}{RGB}{204,153,255}
\definecolor{lightgray}{rgb}{.95,0.95,0.95}
\definecolor{lightgreen}{rgb}{0.3,0.5,0.3}
\definecolor{darkgreen}{rgb}{0.05,0.3,0.05}

\newcommand{\rfield}{\mathbb{R}}

\newcommand{\diag}[1]{\mathop{\rm diag}{#1}}

 \newcommand{\define}{:=}

\newtheorem{myproposition}{Proposition}
\newtheorem{myremark}{Remark}
\newtheorem{myproblemstatement}{Problem Statement}
\newtheorem{mylemma}{Lemma}
\newtheorem{mytheorem}{Theorem}

\newtheorem{mycorollary}{Corollary}

\usepackage{pgfplots}
\pgfplotsset{compat=newest}
\usetikzlibrary{plotmarks}
\usetikzlibrary{arrows.meta}
\usepgfplotslibrary{patchplots}
\usepackage{grffile}

\pgfplotsset{plot coordinates/math parser=false}
\newlength\mywidth
\newlength\myheight
\definecolor{mycolorBL1}{rgb}{0.00000,0.54700,0.54100}%
\definecolor{mycolorBL2}{rgb}{0.85000,0.92500,0.09800}%
\definecolor{mycolorBL3}{rgb}{0.92900,0.69400,0.72500}%
\definecolor{mycolor1}{rgb}{0.00000,0.44700,0.74100}%
\definecolor{mycolor2}{rgb}{0.85000,0.32500,0.09800}%
\definecolor{mycolor3}{rgb}{0.92900,0.69400,0.12500}%
\definecolor{mycolor4}{rgb}{0.89400,0.18400,0.15600}%
\definecolor{mycolor5}{rgb}{0.46600,0.67400,0.18800}%
\definecolor{CGTF}{rgb}{1,0.04700,0.04100}%
\definecolor{SP}{rgb}{0.24220,0.15040,0.66030}%
\definecolor{NewmanF}{rgb}{0.75400,0.59020,0.92180}%
\definecolor{AFG}{rgb}{0.50440,0.79930,0.34800}%
\definecolor{Pott}{rgb}{0.97690,0.98390,0.08050}%

\definecolor{SNMF}{rgb}{0.05000,0.32500,0.89800}%
\definecolor{CGTF1}{rgb}{0.92900,0.69400,0.12500}%
\definecolor{SNMF1}{rgb}{0.49400,0.18400,0.55600}%
\definecolor{CMTF}{rgb}{0.46600,0.67400,0.18800}%
\definecolor{PARAFAC}{rgb}{0.15400,0.59020,0.92180}%
\definecolor{NTF}{rgb}{1,1,0.08400}%
  
\newcommand{\xlabelfontsize}{\small}
\newcommand{\ylabelfontsize}{\small}
\newcommand{\legendfontsize}{\small}
\newcommand{\ticklabelfontsize}{\scriptsize}
\usepackage{bm}

\usepackage{microtype} 
\usepackage{balance}
\usepackage{stfloats}

\usepackage{float} 
\usepackage{hyperref} 
\usepackage{amsmath}
\usepackage{color}
\usepackage{graphicx}
\usepackage{algorithm,algorithmicx,algpseudocode}
\graphicspath{ {images/} }

\newcommand{\spaceforcaption}{0cm}

\newcommand{\change}[1]{{\textcolor{black}{#1}}}
\newif\ifshowtikz
\showtikztrue

\let\oldtikzpicture\tikzpicture
\let\oldendtikzpicture\endtikzpicture

\renewenvironment{tikzpicture}{%
\ifshowtikz\expandafter\oldtikzpicture%
\else\comment%
\fi
}{%
\ifshowtikz\oldendtikzpicture%
\else\endcomment%
\fi
}

\newtheorem{theorem}{Theorem}
\newtheorem{remark}[theorem]{Remark}

\ifCLASSOPTIONcompsoc
  \usepackage[nocompress]{cite}
\else
  \usepackage{cite}
\fi

\begin{document}

\title{ { Coupled Graphs and Tensor Factorization for Recommender Systems  and Community Detection}}
\author{Vassilis~N.~Ioannidis,~\IEEEmembership{Student member,~IEEE,}
        Ahmed~S.~Zamzam,~\IEEEmembership{Student member,~IEEE,}
        Georgios~B.~Giannakis,~\IEEEmembership{Fellow,~IEEE,}
        and~Nicholas~D.~Sidiropoulos,~\IEEEmembership{Fellow,~IEEE}
\IEEEcompsocitemizethanks{\IEEEcompsocthanksitem V. N. Ioannidis, A. S. Zamzam, 
and G. B. Giannakis are with the
ECE Dept. and Digital Tech. Center, Univ. of Minnesota, Mpls,
MN 55455, USA.\protect\\
E-mails:\{ioann006, ahmedz, georgios\}@umn.edu
\IEEEcompsocthanksitem N. D. Sidiropoulos is with the
ECE Dept., Univ. of Virginia, Charlottesville, VA 22903, USA.\protect\\
E-mail: nikos@virginia.edu}
\thanks{
 The work of V. N. Ioannidis and G. B. Giannakis was supported by NSF grants 1442686, 1514056, and 1711471. The work of A. S. Zamzam and N. D. Sidiropoulos was partially supported by NSF grant CIF-1525194.
 V. N. Ioannidis is also supported by the Doctoral Dissertation Fellowship from the University of Minnesota. Preliminary results of this work were presented in~\cite{ioannidis2018imputation,zamzam2016coupled}. A  summary of differences is included in the supplementary material.
}}

\IEEEtitleabstractindextext{%
\begin{abstract}
Joint analysis of data from multiple information repositories facilitates uncovering the underlying structure in heterogeneous datasets. Single and coupled matrix-tensor factorization (CMTF) has been widely used in this context for imputation-based recommendation from ratings, social network, and other user-item data. When this side information is in the form of item-item correlation matrices or graphs, existing CMTF algorithms may fall short. Alleviating current limitations, we introduce a novel model coined coupled graph-tensor factorization (CGTF) that judiciously accounts for graph-related side information. The CGTF model has the potential to overcome practical challenges, such as missing slabs from the tensor and/or missing rows/columns from the correlation matrices. A novel alternating direction method of multipliers (ADMM) is also developed that recovers the nonnegative factors of CGTF. Our algorithm enjoys closed-form updates that result in reduced computational complexity and allow for convergence claims. A novel direction is further explored by employing the interpretable factors to detect graph communities having the tensor as side information. The resulting community detection approach is successful even when some links in the graphs are missing. Results with real data sets corroborate the merits of the proposed methods relative to state-of-the-art competing factorization techniques in providing recommendations and detecting communities. \end{abstract}

}

\maketitle

\IEEEdisplaynontitleabstractindextext

\IEEEpeerreviewmaketitle




\section{Introduction}
Multi-relational data emerge in applications as diverse as social 
networks, recommender systems, biomedical imaging, computer vision and 
communication networks, and are typically modeled
using high-order tensors\cite{sidiro17tensors}.
However, in  many real settings only a subset of the data is observed 
due to application-specific restrictions. 
For example, in recommender systems ratings of new users are missing; 
in social applications individuals may be reluctant to share personal 
information due to privacy concerns; and brain data may contain misses 
due to inadequate spatial resolution. In this context, a task of paramount
importance is to infer unavailable entries given the available data. 

Inference of unavailable tensor data can certainly benefit from
side information that can be available in the form of correlations, 
social interactions, or, biological relations, 
all of which can be captured by a graph~\cite{ioannidis2018kernellearn}. 
In recommender systems for instance, one may benefit from available user-user 
interactions over a social network to impute the missing ratings, and 
also extrapolate (that is predict) profitable recommendations to new costumers.

In addition to graph-aided inference of tensor data, benefits can be 
effected in the opposite direction through  tensor data employed 
to improve graph inference tasks, such as community detection (CD). 
CD amounts to finding 
clusters of vertices densely connected within each cluster 
and scarcely connected across  
clusters~\cite{fortunato2010community}. 
A major challenges emerges here
when some links in the graph are missing due to privacy 
or observation constraints. 
In a social network for example,  not all users 
will provide their social network  connections. 
Additional data organized in a tensor 
can be utilized to improve CD performance and cope
with the missing lnks of the graph.

The present paper develops a novel approach to inference with 
incomplete data by jointly leveraging tensor factorization and
associated graphs. 

\subsection{Related work}
Matrix factorization (MF) techniques have been employed  for
matrix completion  with documented success in user-item
recommender systems \cite{koren2009matrix}.
MF-based techniques assume that the ratings matrix is of low rank, and
hence can be modeled by a reduced number of factors.
Although the  two-relation recommendation model has wide 
applicability, multi-relation data motivate the use of high-dimensional 
tensor models.
Scalable algorithms for nonnegative tensor factorization (TF)
have been pursued \cite{liavas2015parallel}, but do not consider
further structure on tensor modes or any other form of side information.

Side information in the form of matrices sharing factors with a data tensor has been 
investigated in the so-termed 
coupled matrix-tensor factorization (CMTF) 
\cite{ermics2015link,Papalexakis2014turbo,acar2010scalable}. 
Typically, CMTF  adopts
a low-rank model for the tensor to recover the missing entries. 
Misses in both the side information and the tensor were handled 
in~\cite{Papalexakis2014turbo, ermics2015link}, but not with the
use of graph adjacency matrices. 
Using a Bayesian approach, inference relying on tensor 
factorization with low-rank covariance regularization, 
was reported in~\cite{Bazerque-2013}.
Albeit interesting, this approach assumes that the 
similarity matrices are fully observable, which is 
not the case in several applications e.g., social networks.

\subsection{Our Contributions}
Alleviating the limitations of existing
approaches, this paper introduces a novel
factorization model coined coupled graph
and tensor factorization (CGTF) to account 
for the graph structure of side information.
The CGTF factors are estimated via 
a novel algorithm based on the alternating method of 
multipliers (ADMM) to infer missing entries in both
the matrices and the tensor. The CGTF is subsequently
explored to detect communities in the
partially observed coupled graphs.
Specifically, the contribution of this paper is fourfold.
\begin{itemize}
\item[C1.] A novel model is introduced to link multiple repositories
of information bearing data and their correlations in the form 
of high-order tensors and graphs. 
The proposed approach can overcome practical
challenges, such as missing slabs from the tensor and/or missing
rows/columns from the correlation matrices (graph links), 
known as the {\em cold start} problem.

\item[C2.] A novel ADMM algorithm is developed that
features convergence guarantees and low computational
complexity by using closed-form updates.
Our accelerated ADMM solver  
leverages data sparsity~\cite{Papalexakis2014turbo} and 
can easily incorporate other types
of constraints on the latent factors.
\item[C3.] The proposed approach is applied to
recommender systems and markedly improves the 
rating prediction performance. The results in two real 
datasets  corroborate that the novel method is successful
in providing accurate recommendations as well as
recovering missing links in graphs.
\item[C4.] Finally, the proposed coupled factorization 
approach enables detection of communities on graphs by
using the recovered factors. Experiments testify to
the ability of  CGTF to exploit the tensor data for 
CD even when graph links are missing; e.g.,
cold start  problem.
\end{itemize}

\change{
The novel contribution of this work concerning CD is in the coupling between tensor and graph data. Nodes in the recovered communities have similar graph connections and tensor data.  Different than traditional CD methods~\cite{danon2006effect,ronhovde2010local,hespanha2004efficient,arenas2008analysis,kuang2012symmetric,huang2015online,anandkumar2014tensor} 
that find communities given only the graph,  our CGTF finds communities from  the \emph{coupled} tensor and graph data.}

The rest of this paper is organized as follows. 
Sec. II describes the  model 
and the problem formulation.
Sec. III introduces the novel algorithm, and 
Sec. IV deals with the application of CGTF to community detection.
Sec. V demonstrates the effectiveness of the proposed approach in real and 
synthetic data. Finally, Sec. VI summarizes some closing remarks.

Throughout, lower and upper boldface letters are used to denote 
vectors and matrices, respectively. 
The tensors are denoted by underlined upper case boldface symbols. 
For any general matrix $\bf{X}$, ${ \bf{X}}^T $, $ 
{\bf{X}}^{-1} $, $ \Tr({\bf{X}}) $, and $ \diag({\bf{X}}) $ denote 
respectively the transpose,  inverse, trace, and diagonal of $ \bf{X} $. The 
Khatri-Rao and Hadamard products of two matrices $ \bf{X} $ and $ \bf{Y} $ are 
denoted by $ \bf{X} \odot \bf{Y} $ and $  \bf{X} \ast \bf{Y}  $, respectively. 
The operator $ \vectorize(\cdot) $ denotes the vectorization of $ (\cdot) $.

\section{Coupled factorization model} 
Consider a tensor $ \datatensor$ of order $\ordertensor $ and size $ 
\factorsize\factnot{1} \times \factorsize\factnot{2} \times \cdots \times 
\factorsize\factnot{\ordertensor}$. 
An entry of $ \datatensor $ is denoted by $ 
[\datatensor]_{(\tensind\factnot{1},\tensind\factnot{2},\cdots,
\tensind\factnot{N})} $, where  index $ \tensind\factnot{\factind}$ refers to the 
$ \factind $-th mode of the tensor. The focus of this paper is on tensors with  
positive entries that appear in diverse applications such as recommender 
systems, finance, or biology. The  mode-$ k$	matricization of $\datatensor$ 
is denoted by the matrix $ \datamat\factnot{k}$, which arranges the mode-$ k $ 
one-dimensional fibers as columns of the resulting matrix; 
see~\cite{sidiro17tensors} for details. 

Without loss of generality, consider 3-way tensors $ \datatensor\in 
\mathbb{R}^{\factorsize\factnot{1}\times \factorsize\factnot{2} \times 
\factorsize\factnot{3}}_+$.
In many real settings, tensors have low rank and hence can be expressed via the 
well-known 
parallel
factor (PARAFAC) decomposition~\cite{sidiro17tensors} that models a 
rank-$\ranktensor$ tensor as 
\begin{align}[\datatensor]_{(\tensind\factnot{1},\tensind\factnot{2},
\tensind\factnot{3})}=\sum_{\ranktensorind=1}^{\ranktensor}[\factmat\factnot{1}]_ 
{(\tensind\factnot{1},\ranktensorind)}[\factmat\factnot{2}]_
{(\tensind\factnot{2},\ranktensorind)}[\factmat\factnot{3}]_
{(\tensind\factnot{3},\ranktensorind)}+[\errortensor]_{(\tensind\factnot{1},\tensind\factnot{2},
\tensind\factnot{3})}\nonumber
\end{align}
where  
$\{\factmat\factnot{\factind}\in\mathbb{R}_+^{\factorsize\factnot{\factind} 
\times\ranktensor}\}_{\factind=1}^3 $ represent the low-rank factor matrices 
corresponding to the three modes of the tensor, and $\errortensor\in 
\mathbb{R}^{\factorsize\factnot{1}\times \factorsize\factnot{2} \times 
\factorsize\factnot{3}}$ captures model mismatch. The PARAFAC model  is written 
in tensor-matrix form as 	 
\begin{align}
\label{eq:parafacm}
\datatensor= 
\parafacnot{\factmat\factnot{1}}{\factmat\factnot{2}}{\factmat\factnot{3}}+ 
\errortensor
\end{align}
where 
$\parafacnot{\factmat\factnot{1}}{\factmat\factnot{2}}{\factmat\factnot{3}}$ is 
the outer product of these matrices resulting in a tensor. Oftentimes, only a 
subset of entries of $\datatensor$ is observable
due to application-specific constraints such as privacy in social network 
applications; experimental error in the data collection process; or missing 
ratings in recommender systems. Hence,
 we write $\datatensor=\datatensoravail+\datatensormiss$, where 
$\datatensoravail$ contains the available tensor entries and otherwise is zero 
and $\datatensormiss$ holds the missing values and zeros elsewhere. 

\begin{figure}
    \centering
    \includegraphics[width=0.5\textwidth,
    height=0.25\textwidth]{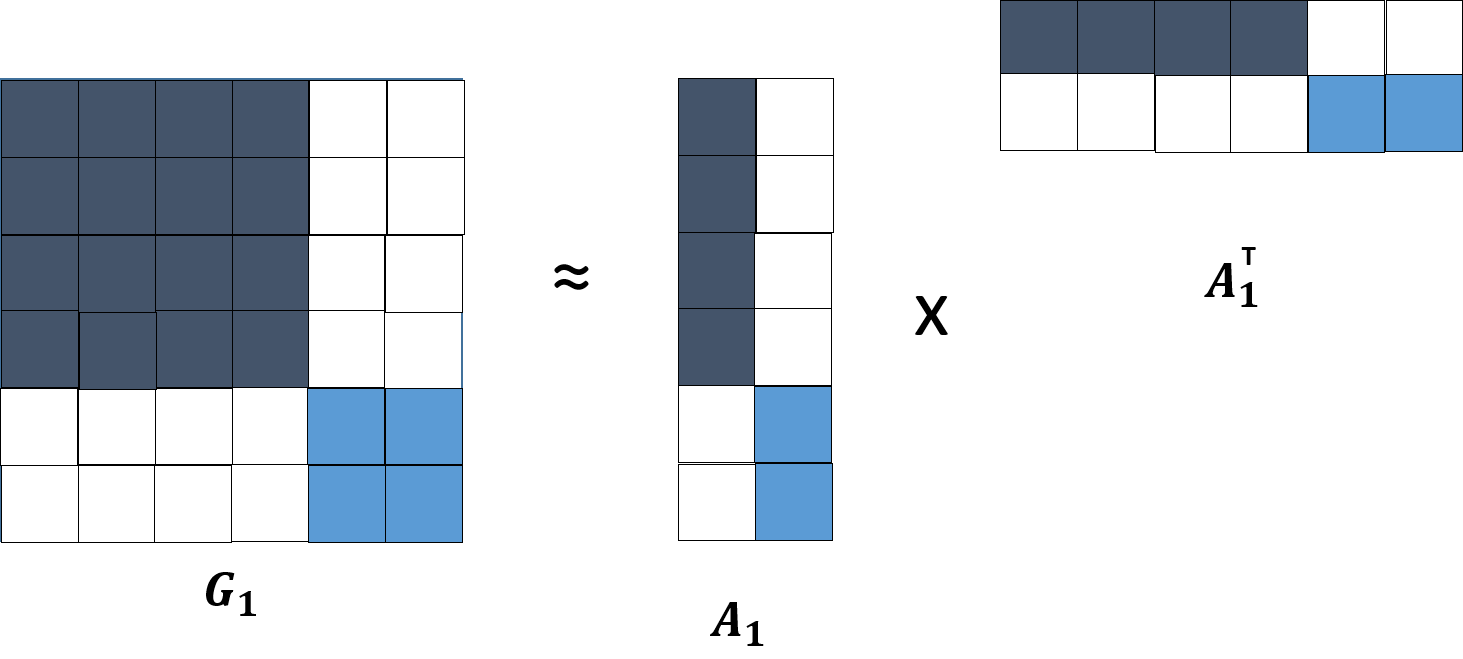}
    \vspace{\spaceforcaption}\caption{Illustration of the SNMF model~\eqref{eq:symnmfm} for $
    \graphmat\factnot{1}$ with $\diagvec\factnot{1}=
    \mathbf{1}$. White cells correspond to
    small-value entries. 
    The rows and columns of
    $\graphmat\factnot{1}$ have been reorganized
    to place nodes in the same community one after the other.}
    \label{fig:nmf_model}
\end{figure}

The tensor entries are also related through 
a set of per-mode graph adjacency or similarity matrices 
$\{\graphmat\factnot{\factind}\in\mathbb{R}_+^
{\factorsize\factnot{\factind} 
\times\factorsize\factnot{\factind}}\}_{\factind=1}^3 $. 
The $ (\tensind,\tensind') $-th 
entry of $ \graphmat\factnot{\factind}$ 
reflects the similarity between the $ \tensind 
$-th and $ \tensind'$-th data items of the 
$\factind$-th tensor mode and thus,  
$\graphmat\factnot{\factind}$ captures 
the connectivity of the corresponding mode-$n$ graph. 
This prior information for the tensor entries  
is well-motivated since network data are available 
across numerous disciplines including sociology, 
biology, neuroscience	 and engineering.   
In these domains, subsets of 
entries (here graph nodes) form communities 
in the sense that they exhibit dense 
intra-connections and sparse inter-connections, 
which are captured by 
$\graphmat\factnot{\factind}$. Such connections are common
in e.g., social networks~\cite{sheikholeslami2017overlapping}, 
where friends tend to form dense 
clusters. We will model this graph-induced side information 
on tensor data using  a symmetric nonnegative 
matrix  factorization (SNMF) 
model~\cite{kuang2012symmetric}, which can efficiently 
provide identifiable factors and recover graph clusters.
Specifically, we advocate the 
following diagonally-scaled SNMF model 
\begin{align}
\label{eq:symnmfm}
\graphmat\factnot{\factind}=\factmat 
\factnot{\factind}\diag{(\diagvec\factnot{\factind})} 
\factmat\factnot{\factind}^\transpose+\errorgraphmat\factnot{\factind},~\factind=1,2,3
\end{align}
where  
$\{\errorgraphmat\factnot{\factind}\in\mathbb{R}^{\factorsize\factnot{\factind}
\times\factorsize\factnot{\factind}}\}_\factind$ capture modeling errors; 
$\{\diagvec\factnot{\factind}\in\mathbb{R}^{\ranktensor\times1}_+
\}_\factind$ weight the factor matrices; and 
$\{\factmat\factnot{\factind}\in\rfield^{\factorsize
\factnot{\factind}\times\ranktensor}\}_\factind$ denote 
factor matrices of rank $\ranktensor<\factorsize
\factnot{\factind}$  that readily reveal communities
in the graphs corresponding to  $\{\graphmat\factnot{\factind}\}_\factind$
\cite{kuang2012symmetric,baingana2016joint}.
Recovering the community of the $\tensind$-th node
in the $\factind$-th
graph is straightforward, by selecting the largest entry 
in the $\tensind$-th row of $\factmat\factnot{\factind}$
\cite{kuang2012symmetric,baingana2016joint};
see Fig.~\ref{fig:nmf_model}.
Unfortunately,  the topologies of $\{\graphmat\factnot{\factind}\}$  
may contain missing entries, 
which can be attributed to privacy concerns in social networks, 
or down-sampling massive 
networks. Hence,  the graph matrices are modeled as
$\graphmat\factnot{\factind}=\graphmatavail\factnot{\factind}+ 
\graphmatmiss\factnot{\factind}$,
 where $\graphmatavail\factnot{\factind}$ contains the available links and 
$\graphmatmiss\factnot{\factind}$ holds the unavailable ones.

\begin{figure}
    \centering
    \includegraphics[width=0.45\textwidth,
    height=0.4\textwidth]{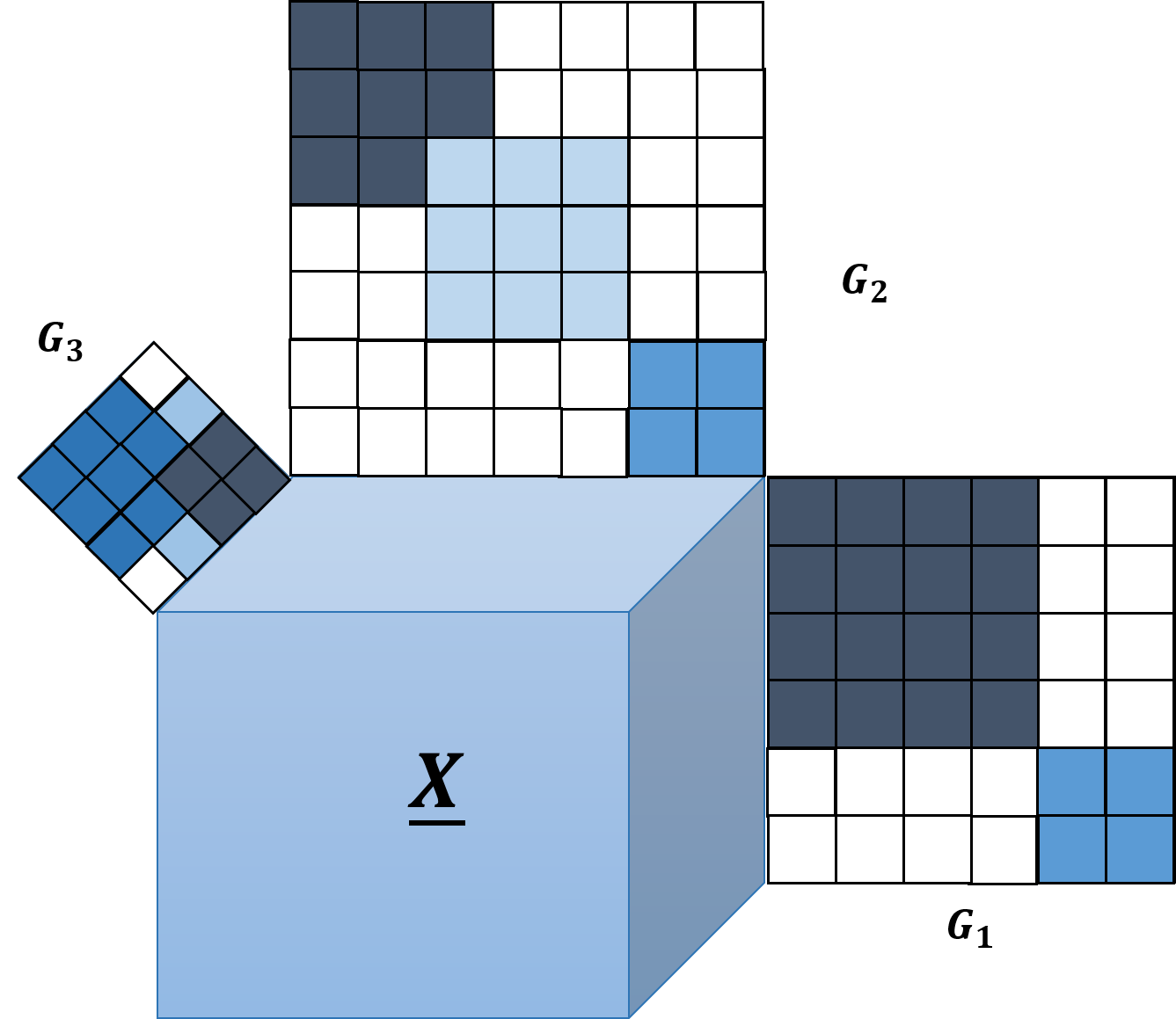}
    \vspace{\spaceforcaption}\caption{Illustration of the tensor and graphs that partake in the CGTF model. The heat maps
    suggest that $\{\graphmat\factnot{\factind}\}_{\factind=1}^3$ exhibit community
    structure.}
    \label{fig:model}
\end{figure}

The factors $\{\factmat\factnot{\factind}\}_\factind$ are 
shared among the tensor and the graph of each corresponding item, 
which justifies the name of the proposed model as coupled graph tensor 
factorization. Whereas classical CMTF approaches model the side information as 
$ \factmat 
	\factnot{\factind} {\bf B}\factnot{\factind}^\transpose$, the novel CGTF 
	captures the graph structure by employing $ \factmat 
	\factnot{\factind}\diag{(\diagvec\factnot{\factind})} 
	\factmat\factnot{\factind}^\transpose $.
Adding the diagonal loading matrices endows the model with the ability to 
adjust the relative weight between the tensor and the side 
information matrices. The novel CGTF model is depicted in Fig.~\ref{fig:model}.

\noindent\textbf{Problem statement.} 
Given $\datatensoravail$ and
$\{\graphmatavail\factnot{\factind}\}_{\factind=1}^3$,
our goal is to estimate 
$\datatensormiss$ and $\{\factmat\factnot{\factind},
\diagvec\factnot{\factind},\graphmatmiss\factnot{\factind}
 \}_{\factind=1}^3$ by 
employing
the CGTF model in \eqref{eq:parafacm} and \eqref{eq:symnmfm}. As a byproduct, the
recovered 
$\{\factmat\factnot{\factind},\diagvec\factnot{\factind}
 \}_{\factind=1}^3$ will be utilized to detect communities.

\section{ Coupled graph tensor factorization}
Given \eqref{eq:parafacm} and \eqref{eq:symnmfm}, this section develops a novel 
algorithm to infer the latent factor matrices and hence estimate $\datatensormiss$ and 
$\{\graphmatmiss\factnot{\factind}\}_{\factind=1}^3$. To this end, 
consider the optimization task
\begin{align}
\label{eq:cgtf}
\underset{\datatensormiss,\{\factmat\factnot{\factind},\diagvec\factnot{\factind},\graphmatmiss\factnot{\factind}
 \}_{\factind=1}^3}{\text{minimize}}~&\|\datatensor
-\parafacnot{\factmat\factnot{1}}{\factmat\factnot{2}}{\factmat\factnot{3}}\|^2_F
 \nonumber\\ +&
\regpar\sum_{\factind=1}^{3}\|\graphmat \factnot{\factind} 
-\factmat\factnot{\factind}\diag{(\diagvec\factnot{\factind})}\factmat\factnot{\factind}^
 \transpose\|^2_F
\nonumber\\
\text{s. t.} ~~~~~~~~~~~~~~~~~~~
&\hspace{-0.4cm}\factmat\factnot{\factind}\ge\mathbf{0},~\diagvec\factnot{\factind}\ge\mathbf{0},\\
&\hspace{-0.4cm}\datatensor=\datatensoravail+\datatensormiss,~
\graphmat\factnot{\factind}=\graphmatavail\factnot{\factind} 
+\graphmatmiss\factnot{\factind},\nonumber\\
&\hspace{-0.4cm}\misstensoperator(\datatensormiss)=\underline{\mathbf{0}},~ 
\missmatoperator{\factind}(\graphmatmiss\factnot{\factind})=\mathbf{0},~\factind=1,2,3\nonumber\nonumber
\end{align}
where $\regpar>0$ tunes the relative importance of the fit between the tensor 
and the graph-induced side information. The first term accounts for the LS 
fitting error of the PARAFAC model~\eqref{eq:parafacm}, and the second sum
of LS costs accounts for the SNMF model~\eqref{eq:symnmfm}. The positivity
constraints  stem from prior knowledge related to the factor and diagonal
matrices. The equality conditions constrain $\datatensor$ and
$\{\graphmat\factnot{\factind}\}_{\factind=1}^3$   to be equal to $\datatensoravail$ 
and $\{\graphmatavail\factnot{\factind}\}_{\factind=1}^3$ at the observed entries 
and to the optimization variables $\datatensormiss$ and 
$\{\graphmatmiss\factnot{\factind}\}_{\factind=1}^3$  otherwise. The operators 
$\misstensoperator$ and $\missmatoperator{\factind}$ force the optimization 
variables  to be zero at the observed entries.

The optimization problem in~\eqref{eq:cgtf} is non-convex due to the trilinear 
terms 
$\parafacnot{\factmat\factnot{1}}{\factmat\factnot{2}}{\factmat\factnot{3}}$ 
and 
$\factmat\factnot{\factind}\diag{(\diagvec\factnot{\factind})}\factmat\factnot{\factind}^
 \transpose$. The next section develops an efficient solver for \eqref{eq:cgtf} 
based on the ADMM~\cite{liavas2015parallel}.

\change{
\begin{remark}
In some applications, a graph $\graphmat_{n}$ may not be available for one or more modes $n$ of the tensor. Hence, before solving~\eqref{eq:cgtf} one may remove the corresponding fitting term $\|\graphmat \factnot{\factind} 
-\factmat\factnot{\factind}\diag{(\diagvec\factnot{\factind})}\factmat\factnot{\factind}^
 \transpose\|^2_F$ and related graph constraints. As a byproduct, our novel framework may utilize the recovered factor $\factmat_{n}$ and obtain a similarity matrix $\graphmat_{n}$~\eqref{eq:symnmfm}.
\end{remark}}

\subsection{ADMM  for CGTF }

First notice that the optimization problem \eqref{eq:cgtf} is even non-convex 
for each $\factmat\factnot{\factind}$ separately due to the product of factor 
matrices in the SMNF 
model. 
This poses an additional challenge to any ADMM algorithm that iteratively 
pursues per block minimizers of the augmented Lagrangian. Hence, we introduce 
$\{\cfactmat\factnot{\factind}\}_\factind$  auxiliary variables 
 and rewrite the SMNF cost as \begin{align}
     \|\graphmat \factnot{\factind} 
-\factmat\factnot{\factind}\diag{(\diagvec\factnot{\factind})}\cfactmat^\transpose
\factnot{\factind}\|_F^2.\end{align} 
Furthermore, to handle the positivity constraints we introduce
\begin{align}
\posconstfun{\mathbf{M}}= \begin{cases} 0,&\text{if $\mathbf{M}\ge 
\mathbf{0}$}\\
\infty,&\text{otherwise}
\end{cases}
\end{align}
and the auxiliary variables 
$\{\pfactmat\factnot{\factind},\pdiagvec\factnot{\factind}\}_\factind$.
Next, we rewrite \eqref{eq:cgtf} to an  equivalent form as 
\begin{align}
\label{eq:cgtfeq}
&\underset{\datatensormiss,\{\factmat\factnot{\factind},\cfactmat\factnot{\factind},
\pfactmat\factnot{\factind},
\atop
\diagvec\factnot{\factind},\pdiagvec\factnot{\factind},\graphmatmiss\factnot{\factind}
 \}_{\factind=1}^3}{\text{minimize}}~\|\datatensor
-\parafacnot{\factmat\factnot{1}}{\factmat\factnot{2}}{\factmat\factnot{3}}\|^2_F
 + \sum_{f=1}^{3}\posconstfun{\pfactmat\factnot{\factind}}  \nonumber\\ &	
+\regpar\sum_{\factind=1}^{3}\|\graphmat \factnot{\factind} 
-\factmat\factnot{\factind}\diag{(\diagvec\factnot{\factind})}\cfactmat^\transpose
 \factnot{\factind}\|^2_F
+
\sum_{f=1}^{3}\posconstfun{\pdiagvec\factnot{\factind}}\nonumber\\
&~~\text{s. t.}~~~~~~~\factmat\factnot{\factind}=\cfactmat\factnot{\factind},~ 
\factmat\factnot{\factind}=\pfactmat\factnot{\factind},~ 
\diagvec\factnot{\factind}=\pdiagvec\factnot{\factind},
\\
&~~~~~~~~~~~~~~~\datatensor=\datatensoravail+\datatensormiss,~
\graphmat\factnot{\factind}=\graphmatavail\factnot{\factind} 
+\graphmatmiss\factnot{\factind},\nonumber\\
&~~~~~~~~~~~~~~~\misstensoperator(\datatensormiss)=\underline{\mathbf{0}},~ 
\missmatoperator{\factind}(\graphmatmiss\factnot{\factind})=\mathbf{0} 
,~\factind=1,2,3\nonumber.
\end{align}
Even though \eqref{eq:cgtfeq} is still non-convex in all the variables, it is 
convex with respect to each block variable separately. Towards deriving an ADMM 
solver, we introduce the dual variables 
$\{\dualadmmmat_{\cfactmat\factnot{\factind}}\in\mathbb{R}^{ 
\factorsize\factnot{\factind}\times\ranktensor}, 
\dualadmmmat_{\pfactmat\factnot{\factind}}\in\mathbb{R}^{ 
\factorsize\factnot{\factind}\times\ranktensor}, 
\dualadmmvec_{\pdiagvec\factnot{\factind}}\in\mathbb{R}^{ 
 \ranktensor\times 1}\}_\factind$ and the penalty 
parameters
$\{\regadmm_{\cfactmat\factnot{\factind}}>0, 
\regadmm_{\pfactmat\factnot{\factind}}>0, 
\regadmm_{\pdiagvec\factnot{\factind}}>0\}_\factind$.  
\begin{figure*}[b]
\begin{align}
\label{eq:lagrang}
\lagragian\big(\datatensormiss,\collectedvar,\{\graphmatmiss\factnot{\factind},
\dualadmmmat_{\cfactmat\factnot{\factind}}, 
\dualadmmmat_{\pfactmat\factnot{\factind}},
\dualadmmvec_{\pdiagvec\factnot{\factind}}\}_
{\factind=1}^3\big) 
:=&\fitfun\big(\datatensormiss,\collectedvar,\{\graphmatmiss\factnot{\factind}
\}_
{\factind=1}^3\big)+
\sum_{f=1}^{3}\big\{
\Tr(\dualadmmmat_{\cfactmat\factnot{\factind}}^\transpose 
(\factmat\factnot{\factind}-\cfactmat\factnot{\factind}) +
\tfrac{\regadmm_{\cfactmat\factnot{\factind}}}{2}\|\factmat\factnot{\factind}
- \cfactmat\factnot{\factind}\|_F^2\\
+\Tr(\dualadmmmat_{\pfactmat\factnot{\factind}}^\transpose 
(\factmat\factnot{\factind}-\pfactmat\factnot{\factind})) 
+&\tfrac{\regadmm_{\pfactmat\factnot{\factind}}}{2}\|\factmat\factnot{\factind}
- \pfactmat\factnot{\factind}\|_F^2+
\dualadmmvec_{\pdiagvec\factnot{\factind}}^\transpose 
(\diagvec\factnot{\factind}-\pdiagvec\factnot{\factind})
+\tfrac{\regadmm_{\pdiagvec\factnot{\factind}}}{2}\|\diagvec\factnot{\factind}
- \pdiagvec\factnot{\factind}\|_F^2
\big\}.\nonumber
\end{align}

\end{figure*}

The augmented Lagrangian is given in \eqref{eq:lagrang}, at the bottom of the 
next page, where $\fitfun(\cdot)$ represents the cost function in \eqref{eq:cgtfeq} 
and we collect all factor variables
in $\collectedvar\define(\{\factmat\factnot{\factind},\cfactmat\factnot{\factind},
\pfactmat\factnot{\factind},\diagvec\factnot{\factind},\pdiagvec\factnot{\factind}
\}_{\factind=1}^3)$.  
For ease of notation no ADMM superscripts will be used in the following 
equations. For brevity, only the ADMM updates for  $\factind=1$ 
will be presented. 

The update for $\factmat\factnot{1}$ can be obtained by 
taking the derivative of $L$ in \eqref{eq:lagrang} with respect to (w.r.t.)
$\factmat\factnot{1}$ and equating it to zero that yields
\begin{subequations}
\label{eq:admmupd}
\begin{align}
\label{eq:facupd}
&\factmatest\factnot{1}  (\katriraofac\factnot{1}^\transpose\katriraofac\factnot{1}+\regpar 
\diagmat\factnot{1}\cfactmat\factnot{1}^\transpose 
\cfactmat\factnot{1}\diagmat\factnot{1} 
+(\regadmm_{\pfactmat\factnot{1}}+\regadmm_{\cfactmat\factnot{1}})\eye_\ranktensor) 
\\&~=\datamat\factnot{1}^\transpose\katriraofac\factnot{1}+\mu\graphmat\factnot{1} 
\cfactmat\factnot{1}\diagmat\factnot{1}+\regadmm_{\cfactmat\factnot{1}}
\cfactmat\factnot{1}+\!\regadmm_{\pfactmat\factnot{1}}\pfactmat\factnot{1}-
 \!\dualadmmmat_{\pfactmat\factnot{1}} - \!\dualadmmmat_{\cfactmat\factnot{1}} 
\nonumber
\end{align}
where $\katriraofac\factnot{1}:=\factmat\factnot{3} \odot\factmat\factnot{2}$,  
and $\diagmat\factnot{1}:=\diag{(\diagvec\factnot{1})}$. The update for 
$\diagvec\factnot{1}$ can be obtained likewise as
\begin{align}
\label{eq:diagupd}
&((\cfactmat\factnot{1}\odot\factmat\factnot{1})^\transpose 
({\regpar}\cfactmat\factnot{1}\odot\factmat\factnot{1})+\regadmm_{\pdiagvec\factnot{1}}\eye_\ranktensor)
 \diagvecest\factnot{1}
\\&~={\regpar}(\cfactmat\factnot{1}\odot\factmat\factnot{1})^\transpose\graphvec\factnot{1}
 +{\regadmm_{\pdiagvec\factnot{1}}}\pdiagvec\factnot{1}-
\dualadmmvec_{\pdiagvec\factnot{1}}\nonumber.
\end{align}
where $\graphvec\factnot{\factind}\define\vectorize(\graphmat\factnot{\factind})$.
Accordingly, the update for the $\cfactmat\factnot{1}$ is given by
\begin{align}
\label{eq:cfacupd}
&\cfactmatest\factnot{1} (\regpar 
\diagmat\factnot{1}\factmat\factnot{1}^\transpose 
\factmat\factnot{1}\diagmat\factnot{1} +\regadmm_{\cfactmat\factnot{1}}\eye_\ranktensor) 
\\&~=\mu\graphmat\factnot{1}^\transpose 
\factmat\factnot{1}\diagmat\factnot{1}+\regadmm_{\cfactmat\factnot{1}}\factmat\factnot{1}+
 \!\dualadmmmat_{\cfactmat\factnot{1}} \nonumber.
\end{align}
The auxiliary variables $\pfactmat\factnot{1},\pdiagvec\factnot{1}$ are updated 
by  projecting to the nonnegative orthant as follows
\begin{align}
\label{eq:pfacupd}
\pfactmatest\factnot{1}=&\bigg(\factmat\factnot{1}+\frac{1}{ 
\regadmm_{\pfactmat\factnot{1}}}\dualadmmmat_{\pfactmat\factnot{1}}\bigg)_+,
\nonumber\\
\pdiagvecest\factnot{1}=&\bigg(\diagvec\factnot{1}+\frac{1}{ 
\regadmm_{\pdiagvec\factnot{1}}}\dualadmmvec_{\pdiagvec\factnot{1}}\bigg)_+ .
\end{align}
Using the estimated factors 
$\{
\factmatest\factnot{\factind}\}_\factind$ the updates for the missing tensor 
elements are given by 
\begin{align}
\label{eq:misstensupd}
\datatensormissest=
\misstensoperator(\parafacnot{\factmatest\factnot{1}}{\factmatest\factnot{2}} 
{\factmatest\factnot{3}}).
\end{align}
Similarly, the missing entries in $\graphmat\factnot{1}$ can be obtained by
\begin{align}
\label{eq:missgraphupd}
\graphmatmissest\factnot{1}=
\missmatoperator{1}(\factmatest\factnot{1}\diag{(\diagvecest\factnot{1})} 
\cfactmatest^\transpose \factnot{1}).
\end{align}

Finally, the updates for the Lagrange multipliers are
\begin{align}
\dualadmmmat_{\cfactmat\factnot{1}}=&\dualadmmmat_{\cfactmat\factnot{1}}+
\regadmm_{\cfactmat\factnot{1}}(\factmat\factnot{1}-\cfactmat\factnot{1})\nonumber\\
\dualadmmmat_{\pfactmat\factnot{1}}=&\dualadmmmat_{\pfactmat\factnot{1}}+
\regadmm_{\pfactmat\factnot{1}}(\factmat\factnot{1}-\pfactmat\factnot{1})\nonumber\\
\dualadmmvec_{\pdiagvec\factnot{1}}=& \dualadmmvec_{\pdiagvec\factnot{1}}+
\regadmm_{\pdiagvec\factnot{1}}(\diagvec\factnot{1}-\pdiagvec\factnot{1}).
\label{eq:lagrangupd}
\end{align}
\end{subequations}

The steps of our CGTF algorithm are listed in Algorithm 1.
Since \eqref{eq:cgtfeq} is a non-convex problem, a judicious initialization of 
$\{\factmat\factnot{\factind}\}_\factind$ is required. Towards that end, we 
adopt an efficient algorithm for SNMF, see~\cite{huang2014putting}, to 
initialize the 
factor matrices using only the available elements in the corresponding graphs 
$\{\graphmatavail\factnot{\factind}\}$, while  
$\{\diagvec\factnot{\factind}\}$ are 
initialized as all-ones vectors. Since SNMF is unique under certain conditions, the 
initialization is 
likely to be a good one~\cite{huang2014putting}. The ADMM algorithm stops 
when 
the primal residuals
and the dual feasibility residuals are sufficiently small. Even though $\{\pfactmat\factnot{\factind},
\pdiagvec\factnot{\factind}\}_{n}$ are by construction 
non-negative, 
$\{\factmat\factnot{\factind},\cfactmat\factnot{\factind},
\diagvec\factnot{\factind}\}_{n}$ are not necessarily non-negative,
but they become so upon convergence.

The advantage of introducing the auxiliary variables is threefold.
First, by employing $\cfactmat\factnot{\factind}$, we bypass solving the 
non-convex 
SNMF 
that would require a costly iterative algorithm per factor update.
Second,  by introducing $\{\pfactmat\factnot{\factind}, 
\pdiagvec\factnot{\factind}\}$, we 
avoid the solution to a constrained optimization problem, resulting in 
a more computationally affordable update compared to constrained 
least-squares based algorithms.
In a nutshell, our novel reformulation allows for 
closed-form updates per step of the ADMM solver. 
Lastly, the closed-form updates allow us to make 
convergence claims to a stationary point 
of \eqref{eq:cgtfeq} in Sec. \ref{sec:cov_gur}.

\begin{remark}The era of data science brings opportunities for adversaries that aim to corrupt the data, e.g., recommendation data may be corrupted by malicious users that provide fake ratings, or social networks may contain spamming users. The CGTF model can be extended to account for anomalies in the graph links and the tensor data. Specifically, consider the robust CGTF (R-CGTF) as $\datatensor=\parafacnot{\factmat\factnot{1}}{\factmat\factnot{2}}{\factmat\factnot{3}}+\anomaltensor +\errortensor$ and $\graphmat\factnot{\factind}=\factmat \factnot{\factind}\diag{(\diagvec\factnot{\factind})} \factmat\factnot{\factind}^\transpose +\anomalgraphmat\factnot{\factind}+\errorgraphmat\factnot{\factind}$ for the tensor and the graph matrices respectively. The variables $\anomaltensor\in\rfield^{\factorsize\factnot{1}\times\factorsize\factnot{2} \times \factorsize\factnot{3}}$ and $\{\anomalgraphmat\factnot{\factind}\in\rfield^{\factorsize\factnot{\factind}\times\ranktensor}\}_{\factind}$ model the anomalies in the tensor and graphs that should occur infrequently, and hence most entries of $\anomaltensor$ and $\{\anomalgraphmat\factnot{\factind}\}_\factind$ are zero. Hence, the optimization \eqref{eq:cgtf} and the ADMM solver can be readily modified to obtain sparse estimates of $\anomaltensor$ and $\{\anomalgraphmat\factnot{\factind}\}_\factind$ as well; see e.g., \cite{mardani2013dynamic} and \cite{baingana2016joint}.\end{remark}

\subsection{Convergence}\label{sec:cov_gur}
Here, convergence of Algorithm 1 is examined when
all the measurements are available $\{\graphmatavail\factnot{\factind}=\graphmat\factnot{\factind}\}_\factind$, and $\datatensoravail=\datatensor$, the extension for the case with misses is straightforward~\cite{xu2012alternating}. 

A point $\collectedvar\define(\{\factmat\factnot{\factind},\cfactmat\factnot{\factind},
\pfactmat\factnot{\factind},\diagvec\factnot{\factind},\pdiagvec\factnot{\factind}
\}_{\factind=1}^3)$ satisfies the Karush-Kuhn-Tucker (KKT) conditions for problem \eqref{eq:cgtfeq} if there exist dual variables 
$\collecteddualvar\define(\{\dualadmmmat_{\cfactmat\factnot{\factind}}, \dualadmmmat_{\pfactmat\factnot{\factind}}, \dualadmmvec_{\pdiagvec\factnot{\factind}}\}^3_{\factind=1})$ such that 
	\begin{align}
\label{eq:kkt}
	(\datamat\factnot{\factind}-\factmat\factnot{\factind} \katriraofac\factnot{\factind}^\transpose)\katriraofac\factnot{\factind} +\regpar (\graphmat\factnot{\factind}-
	\factmat\factnot{\factind}\diagmat\factnot{\factind} \cfactmat\factnot{\factind}^\transpose) 
	\cfactmat\factnot{\factind}&\diagmat\factnot{\factind}\\
	-\!\dualadmmmat_{\pfactmat\factnot{\factind}} - \!\dualadmmmat_{\cfactmat\factnot{\factind}} 
	&=\mathbf{0}\nonumber\\
	{\regpar}(\cfactmat\factnot{\factind}\odot\factmat\factnot{\factind})^\transpose (\graphvec\factnot{\factind}
	-	\cfactmat\factnot{\factind}\odot\factmat\factnot{\factind}\diagvec\factnot{\factind})-
	\dualadmmvec_{\pdiagvec\factnot{\factind}}&=\mathbf{0}\nonumber\\
	\regpar (\graphmat\factnot{\factind}-
	\cfactmat\factnot{\factind}\diagmat\factnot{\factind} \factmat\factnot{\factind}^\transpose) 
	\factmat\factnot{\factind}\diagmat\factnot{\factind}
- \!\dualadmmmat_{\cfactmat\factnot{\factind}} 
	&=\mathbf{0}\nonumber\\
	\factmat\factnot{\factind}-\cfactmat\factnot{\factind}=\mathbf{0},~~~~~
	\factmat\factnot{\factind}-\pfactmat\factnot{\factind}=\mathbf{0},~~~~~
	\diagvec\factnot{\factind}-\pdiagvec\factnot{\factind}&=\mathbf{0}\nonumber\\
	\dualadmmmat_{\pfactmat\factnot{\factind}}\le\mathbf{0}\ge\pfactmat\factnot{\factind},~~~~~~
	\dualadmmvec_{\pdiagvec\factnot{\factind}}\le\mathbf{0}&\ge\pdiagvec\factnot{\factind}\nonumber\\
	\dualadmmmat_{\pfactmat\factnot{\factind}}\elementwiseproduct \pfactmat\factnot{\factind}=\mathbf{0},~~~~
	\dualadmmvec_{\pdiagvec\factnot{\factind}}\elementwiseproduct \pdiagvec\factnot{\factind}=\mathbf{0},~~~~~\factind=1,2,3.&\nonumber
	\end{align}
\begin{myproposition}
\label{prop:conv}
Let $\{\collectedvar\admmiternot\admmiter,
\collecteddualvar\admmiternot\admmiter\}_{\admmiter}$ 
be a sequence generated by Algorithm 1. If the sequence of dual variables
$\{\collecteddualvar\admmiternot\admmiter\}_\admmiter$  is bounded and satisfies 
\begin{align}
\sum_{l=0}^{\infty}\|\collecteddualvar\admmiternot
{\admmiter+1}-\collecteddualvar\admmiternot\admmiter
\|^2_F
<\infty\label{eq:thassump}
\end{align}
then any accumulation point of $\{\collectedvar^l\}_l$
satisfies the KKT conditions of \eqref{eq:cgtfeq}.
Hence, any accumulation point of
$\{\{\factmat\factnot{\factind}^l,\diagvec\factnot{\factind}^l
\}_{\factind=1}^3\}_l$ satisfies the KKT conditions for problem \eqref{eq:cgtf}.
\end{myproposition}
\begin{IEEEproof}
See Sec. \ref{proof:prop}.
\end{IEEEproof}
Proposition \ref{prop:conv} suggests that upon convergence 
of the dual variables $\{\collecteddualvar\admmiternot\admmiter\}_\admmiter$,  the
sequence $\{\collectedvar^l\}_l$ reaches a KKT point.
Note that the closed-form updates of 
Algorithm~\ref{algo:cgtf} are instrumental in
establishing the convergence claim.
Empirical convergence with numerical tests
is provided in Sec. \ref{sec:sims}.

%

%

\begin{algorithm}[h]                
\caption{ADMM for CGTF}
\label{algo:cgtf}    
\vspace{0.2cm}
\begin{minipage}{40cm}
	\indent\textbf{Input:} $\datatensoravail$ and 
	$\{\graphmatavail\factnot{\factind}\}_{\factind=1}^3$
	\vspace{0.2cm}
	\begin{algorithmic}[1]
		\State\emph{Initialization:} SNMF for 
		$\{\factmat\factnot{\factind}\}_\factind$ using \cite{huang2014putting}.
		\vspace{0.1cm}
		\State 	\textbf{while} {iterates not converge}
		\textbf{do}\hspace{0.1cm}
		\State \hspace{0.5cm}Update $\factmatest\factnot{\factind}$ using 
		\eqref{eq:facupd}.
		\State \hspace{0.5cm}Update $\diagvecest\factnot{\factind}$ using 
		\eqref{eq:diagupd}.
		\State \hspace{0.5cm}Update $\cfactmatest\factnot{\factind}$ using 
		\eqref{eq:cfacupd}.
		\State \hspace{0.5cm}Update 
		$\{\pfactmatest\factnot{\factind},\pdiagvecest\factnot{\factind}\}$ using 
		\eqref{eq:pfacupd}.
		\State \hspace{0.5cm}Update $\datatensormissest$ using 
		\eqref{eq:misstensupd}.
		\State \hspace{0.5cm}Update $	\graphmatmissest\factnot{\factind}$ using 
		\eqref{eq:missgraphupd}.
			\State \hspace{0.5cm}Update Lagrange multipliers using 
			\eqref{eq:lagrangupd}.
		\State\textbf{end while}
	\end{algorithmic}
	\vspace{0.2cm}
	\indent\textbf{Output:} 
	$\datatensormissest,\{\factmatest\factnot{\factind},\diagvecest\factnot{\factind},
	\graphmatmissest\factnot{\factind}\}_\factind$ 
\end{minipage}
\end{algorithm}

\section{Community detection via CGTF}
\label{sec:com_det}
A task of major practical importance in network science is the identification of
groups of vertices or communities that are more densely connected to each 
other than to the rest of the nodes in the network. 
Community detection unveils the structure of the network and 
facilitates a number of applications. For example, clustering 
web clients improves the performance of web services, identifying
communities among customers leads to accurate recommendations,
or grouping proteins based on their dependencies enables the 
development of targeted drugs~\cite{fortunato2010community}. 
This section exemplifies how the novel CGTF can recover the communities 
in graphs even when some links are missing; what is known as the
cold start  problem.

Community detection methods aim to learn for each node
$\tensind\in\{1,\ldots,\factorsize\factnot{\factind}\}$ of 
$\graphmat\factnot{\factind}$ a mapping to a cluster assignment  
$\assignment\factnot{\factind,\tensind}\in \{1,\ldots,\commnbr\factnot{\factind}\}$, 
where $\commnbr\factnot{\factind}$
is the number of communities in the $\factind$-th graph. 
Collecting all the nodal assignments,
one seeks an $\factorsize\factnot{\factind}\times1$ vector  
$\assignmentvec\factnot{\factind}\define[\assignment\factnot{\factind,1},\ldots,
\assignment\factnot{\factind,\factorsize\factnot{\factind}}]^\transpose$.

If $\commnbr\factnot{\factind}$
is not known a priori, the recovered factor $\factmat\factnot{\factind}$
can be directly utilized to provide a community assignment. First, we scale $\factmat\factnot{\factind}$
to account 
for the weighting vector  $\factmatforcom
\factnot{\factind}\define\factmat\factnot{\factind}	
\diag(\sqrt{\diagvec\factnot{\factind}})$. 
The largest entry in each row of $\factmatforcom\factnot{\factind}$ indicates 
clustering assignments~\cite{kuang2012symmetric}.
Specifically, we estimate the community assignment of a node $\tensind$ as
\begin{align}
\label{eq:comass}
    \estassignment\factnot{\factind,\tensind}=
		\underset{\ranktensorind=1,\ldots,\ranktensor}{\arg\max}~[\factmatforcom
		\factnot{\factind}]_{(\tensind,\ranktensorind)}
\end{align}
and $\estassignmentvec\factnot{\factind}\define
[\estassignment\factnot{\factind,1},\ldots,
\estassignment\factnot{\factind,\factorsize\factnot
{\factind}}]^\transpose$ is the estimated assignment vector.  
Hence, in lieu of prior information about $\commnbr
\factnot{\factind}$ we implicitly assume that 
$\commnbr\factnot{\factind}=\ranktensor$ for $\factind=1,2,3$.

Oftentimes, in CD problems $\commnbr\factnot{\factind}$ is available.
If $\commnbr\factnot{\factind}\ne\ranktensor$ one cannot apply
directly  \eqref{eq:comass} to recover the communities.
In such a case, we regard $\factmatforcom\factnot{\factind}$ as a representation
of $\graphmat\factnot{\factind}$ in a latent space of lower dimension. Hence, we apply the
celebrated k-means algorithm \cite{hartigan1979algorithm} 
obtain
\begin{align}
\label{eq:kmeans}
    	\estassignmentvec\factnot{\factind}=k\text{-means}(\factmatforcom
		\factnot{\factind},\commnbr\factnot{\factind})
\end{align}
The community assignment procedure is summarized in Algorithm \ref{algo:comm_cgtf}.
Note that the discussed method amounts to a hard community assignment in the
sense that each node is assigned to exactly one community. Nonetheless, the factors
can be utilized to perform soft community assignment, where one node may belong
to more than one communities. If the rows of
$\factmatforcom\factnot{\factind}$ are normalized to sum to 1, $[\factmatforcom
\factnot{\factind}]_{(\tensind,\ranktensorind)}$ can be interpreted as
the probability of the $\tensind$-th node belonging to the $\ranktensorind$-th
community.

\begin{algorithm}[h]                
\caption{Community detection via CGTF}
\label{algo:comm_cgtf}    
\vspace{0.2cm}
\begin{minipage}{40cm}
	\indent\textbf{Input:} $\datatensoravail$, 
	$\{\graphmatavail\factnot{\factind}\}_{\factind=1}^3$, and $\{\commnbr\factnot{\factind}\}_{\factind=1}^3$ 
	\vspace{0.2cm}
	\begin{algorithmic}[1]
		\State\emph{Initialization:} Algorithm 1 for 
		$\{\factmat\factnot{\factind},\diagvec\factnot{\factind}\}^3_{\factind
		=1}$.
		\State \textbf{for} $\factind=1,2,3$
		\State\hspace{0.2cm}$\factmatforcom
		\factnot{\factind}\define\factmat\factnot{\factind}
		\diag(\sqrt{\diagvec\factnot{\factind}})$
		\State \hspace{0.2cm}\textbf{if} {$\commnbr\factnot{\factind}=\ranktensor$}
		\textbf{do}
		\State\hspace{0.7cm}Compute $\estassignmentvec\factnot{\factind}$
		using \eqref{eq:comass}
		\State\hspace{0.2cm}\textbf{else}
		\State\hspace{0.7cm}Compute $\estassignmentvec\factnot{\factind}$ using 
		\eqref{eq:kmeans}
	\end{algorithmic}
	\vspace{0.2cm}
	\indent\textbf{Output:} 
	$\{\estassignmentvec\factnot{\factind}\}_\factind$ 
\end{minipage}
\end{algorithm}

\subsection{Community detection evaluation}
\label{sec:com_det_eval}
For a graph of $\factorsize$ nodes and graph adjacency matrix $\graphmat$, 
we define the cover set $\cover\define\{\comunitycover_{\comind}\}_{
\comind=1}^\commnbr$ where $\comunitycover_{\comind}$ 
contains all the nodes that belong to community 
$\comind$ as captured by the assignment vector $\assignmentvec$, i.e.,  
$\comunitycover_{\comind}\define\{\tensind| \text{s.t. }
\assignment\factnot{\tensind}=\comind\}$. The estimated cover set is defined as $\estcover$
that uses $\estassignmentvec$ from Algorithm \ref{algo:comm_cgtf}.

For networks with ground truth communities,  we employ the 
normalized mutual information (NMI) metric \cite{danon2005comparing} to 
evaluate the recovered communities, $\estassignmentvec$.  
The NMI takes values between 0 and 1 and is defined  as
\begin{align}
    \text{NMI}\define\frac{2\mutualinfo{\cover}{\estcover}}
    {\entropy{\cover}+\entropy{\estcover}}
\end{align}
where H denotes the entropy ($|\cal C|$ is the cardinality of $\cal C$) 
\begin{align}
    \entropy{\cover}\define-\sum_{\comind=1}^
    \commnbr\frac{|\comunitycover_{\comind}|}{\factorsize}
\log\frac{|\comunitycover_{\comind}|}{\factorsize}
\end{align}
and $\mutualinfo{\cover}{\estcover}$ 
stands for the mutual information (MI) between
$\cover$ and $\estcover$ defined as
\begin{align}
   \mutualinfo{\cover}{\estcover}\define
   \sum_{\comind=1}^{\commnbr}\sum_{\comind'=1}^{
  \hat{\commnbr}}
   \frac{|\estcomunitycover_{\comind'}\cap\comunitycover_{\comind}|}{\factorsize}
\log\frac{|\estcomunitycover_{\comind'}\cap\comunitycover_{\comind}|\factorsize}
{|\estcomunitycover_{\comind'}||\comunitycover_{\comind}|}.
\end{align}
Whereas MI encodes how similar 
two community cover sets are, the entropy measures the level of uncertainty 
in each cover set individually; see e.g.,
\cite{fortunato2010community}. 
For successful  clustering algorithms,  the resulting NMI is 
close to 1, and otherwise 0.

On the other hand, to evaluate the quality of a recovered community
$\estcomunitycover$ even without ground-truth 
community labels, the conductance $\conductance{\estcomunitycover}$ is traditionally employed 
\cite{bollobas2013modern}
\begin{align}
    \conductance{\estcomunitycover}\define\frac{\sum_{\tensind\in\estcomunitycover}
    \sum_{
    \tensind'\notin\estcomunitycover}[\graphmat]_{(\tensind,\tensind')}}
    {\min(\text{vol}
    (\estcomunitycover),\text{vol}(\estcomunitycover^c))}
\end{align}
where
\begin{align}
    \text{vol}(\estcomunitycover)
    \define\sum_{\tensind\in\estcomunitycover}\sum_{
    \tensind'=1}^\factorsize[\graphmat]_{(\tensind,\tensind')}
\end{align}
and the set $\estcomunitycover^c$ contains all nodes in the graph
not in
$\estcomunitycover $. For successful CD, the 
connections among nodes in $\estcomunitycover$  are dense and
otherwise sparse 
that leads to small scores of $\conductance{\estcomunitycover}$.

A  metric that summarizes the conductance across communities
$\{\estcomunitycover_\comind\}_\comind\in\estcover$ is the 
so-termed coverage
\begin{align}
\label{eq:cov}
    \coverage{\estcover,\covscalar}\define\frac{1}{\factorsize}\big|\!\!\underset
    {\conductance{\estcomunitycover_\comind}<\covscalar}
    {\bigcup}
    \estcomunitycover_\comind\big|,~~\{
    \estcomunitycover_\comind\}_\comind\in\estcover
\end{align}
where $\covscalar\in[0,1]$ is a suitable threshold. 
The coverage gives the portion of nodes that belong to communities with 
conductance less than $\covscalar$ and since low conductance scores 
correspond to more cohesive communities, large values of coverage for small
thresholds are desirable.

\section{Experimental evaluation}
\label{sec:sims}
This section evaluates the performance of the proposed CGTF on synthetic and 
real data. The approaches compared 
 include the CANDECOMP/PARAFAC Weighted OPTimization (PARAFAC) 
algorithm~\cite{acar2010scalable}; the nonnegative tensor factorization 
(NTF) implemented as in~\cite{andersson2000n}; 
and the CMTF \cite{ermics2015link}.  The algorithms were initialized using 
the proposed  SNMF scheme, which enhances the performance of all methods.
Unless stated otherwise, the following
parameters were selected for CGTF:  $\{\regadmm_{\cfactmat\factnot{\factind}}=100,
		\regadmm_{\pfactmat\factnot{\factind}}=100,
		\regadmm_{\pdiagvec\factnot{\factind}}=100\}_\factind, \regpar=1$.\footnote{\change{The ADMM implementation of the proposed CGTF method can be found in \underline{\url{https://github.com/bioannidis/Coupled_tensors_graphs}}}}
\subsection{Tensor Imputation}
Synthetic tensor data $\datatensor\in\mathbb{R}^{350\times 350\times 30 }$
with  $\ranktensor=4 $ was generated according to the PARAFAC 
model~\eqref{eq:parafacm}, where the true factors 
$\{\factmat\factnot{\factind}\}_{\factind=1}^3$ are drawn from a 
uniform distribution. Matrices 
$\{\graphmat\factnot{\factind}\}_{\factind=1}^3$ were generated using the SMNF 
\eqref{eq:symnmfm}.

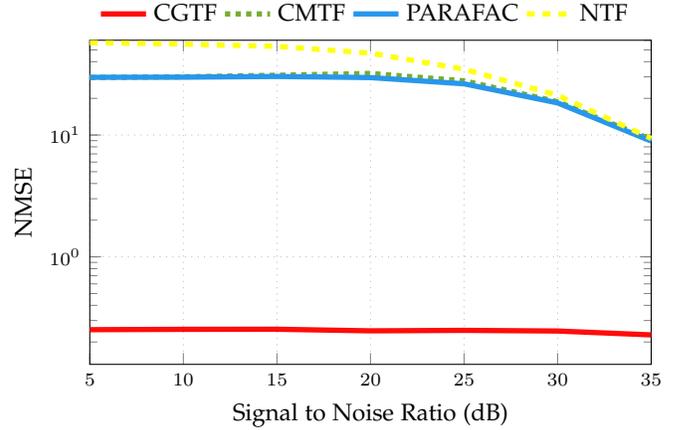
\begin{figure}[h]\centering{
%
%
%
\begin{tikzpicture}

\begin{axis}[%
width=\mywidth,
height=0.987\myheight,
at={(0\mywidth,0\myheight)},
scale only axis,
xmin=5,
xmax=35,
xlabel style={font=\xlabelfontsize},
xlabel={Signal to Noise Ratio (dB)},
ymin=0,
ymax=60,
ylabel style={font=\ylabelfontsize},
ylabel={NMSE},
ymode=log,
yticklabel style={scaled ticks=false,
	/pgf/number format/fixed,
	/pgf/number format/precision=3},
axis background/.style={fill=white},
xmajorgrids,
ymajorgrids,
grid style={dotted},ticklabel style={font=\ticklabelfontsize},
legend columns=4,
legend style={
	at={(0,1.015)}, 
	anchor=south west, legend cell align=left, align=left, draw=none
	, font=\legendfontsize}
]
\addplot [color=CGTF, line width=2.0pt]
  table[row sep=crcr]{%
5	0.252356926774771\\
10	0.254028898577378\\
15	0.254434451852369\\
20	0.246922149964569\\
25	0.249134281367055\\
30	0.246048326411262\\
35	0.228747594257108\\
};
\addlegendentry{CGTF}

\addplot [color=CMTF, dotted, line width=2.0pt]
  table[row sep=crcr]{%
5	29.6220706308717\\
10	30.0590964192704\\
15	30.9502350956693\\
20	32.0273608797385\\
25	27.8518993235934\\
30	18.7619535327313\\
35	9.21029650305723\\
};
\addlegendentry{CMTF}

\addplot [color=PARAFAC, line width=2.0pt]
  table[row sep=crcr]{%
5	29.7788948273644\\
10	29.8509071609964\\
15	30.2592581276236\\
20	29.6191703775326\\
25	26.3303579782052\\
30	18.3566597349053\\
35	8.91983049200881\\
};
\addlegendentry{PARAFAC}

\addplot [color=NTF, dashed, line width=2.0pt]
  table[row sep=crcr]{%
5	57.3423079179448\\
10	55.7705685474406\\
15	53.4205717795496\\
20	47.0332938917996\\
25	34.6832502822569\\
30	21.1945785218184\\
35	9.37955259003011\\
};
\addlegendentry{NTF}

\end{axis}
\end{tikzpicture}%
}
	\vspace{\spaceforcaption}\caption{Tensor imputation performance based on 
	NMSE.}
	\label{fig:nmsedifmodel}
\end{figure}

To evaluate the performance of the various factorization algorithms, the 
entries of $\datatensor$ were corrupted with i.i.d. Gaussian noise.
Fig. 
\ref{fig:nmsedifmodel} depicts 
the normalized mean squared error NMSE$:=
\sum_{i_3=1}^{I_3}\|\hat{\datatensor}(:,:,i_3)-
\datatensor (:,:,i_3)\|_F^2/  
\sum_{i_3=1}^{I_3} \|\datatensor(:,:,i_3)\|_F^2$ against the signal-to-noise ratio (SNR)
of the tensor data. . The novel CGTF exploits 
the  graph adjacency
matrices and achieves superior performance relative to the competing methods.

Furthermore, the convergence of the proposed approach
is evaluated. Fig.~\ref{fig:conv} testifies to
the theoretical convergence results established in
Prop. 1. 
\begin{figure}[h]\centering{\input{figs/conv.tex}}\vspace{\spaceforcaption}\caption{Convergence of  ADMM iterates	$\{\|\collectedvar\admmiternot{\admmiter}-\collectedvar\admmiternot{\admmiter-1}\|_F^2,\|\collecteddualvar\admmiternot{\admmiter}-\collecteddualvar\admmiternot{\admmiter-1}\|_F^2\}_{\admmiter}$, and 
$\|\{\factmat\factnot{\factind}\admmiternot{\admmiter}-\pfactmat\factnot{\factind}\admmiternot{\admmiter}\|_F^2,\|\factmat\factnot{\factind}\admmiternot{\admmiter}-\cfactmat\factnot{\factind}\admmiternot{\admmiter}\|_F^2,\|\diagvec\factnot{\factind}\admmiternot{\admmiter}-\pdiagvec\factnot{\factind}\admmiternot{\admmiter}\|_F^2\}_{\admmiter}$.}\label{fig:conv}\end{figure}

\subsection{Community detection}
\begin{figure*}[h]
    \centering
    \begin{subfigure}[b]{0.32\textwidth}
        \centering
        \includegraphics[height=2.5in]{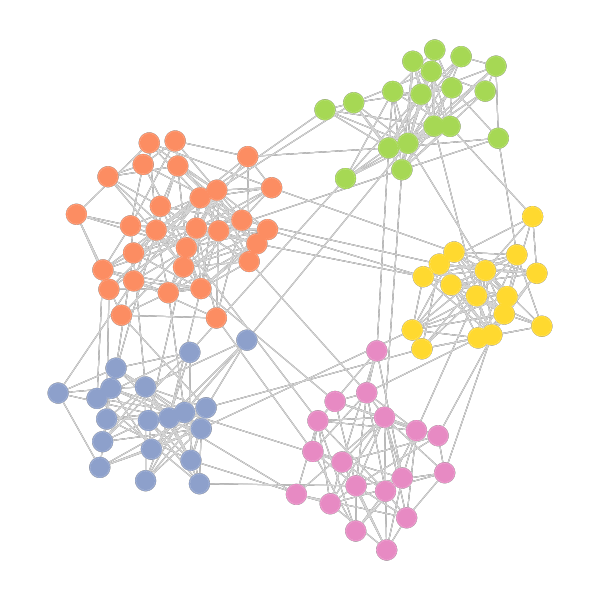}
    \end{subfigure}%
    ~ 
    \begin{subfigure}[b]{0.32\textwidth}
        \centering
        \includegraphics[height=2.5in]{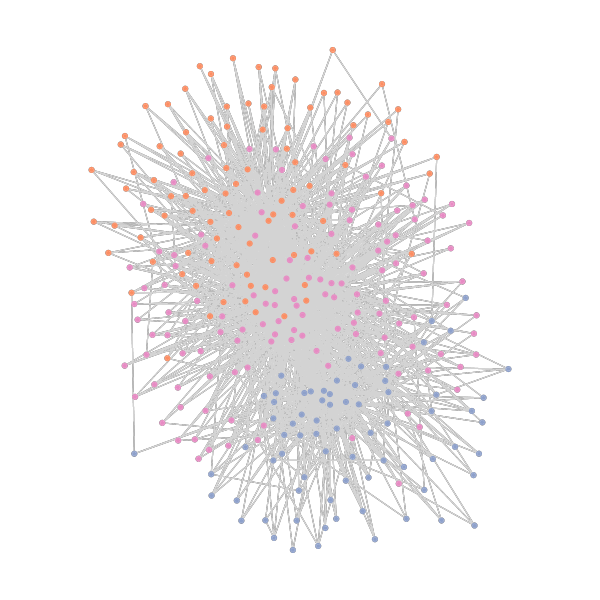}
    \end{subfigure}
        ~ 
    \begin{subfigure}[b]{0.32\textwidth}
        \centering
        \includegraphics[height=2.5in]{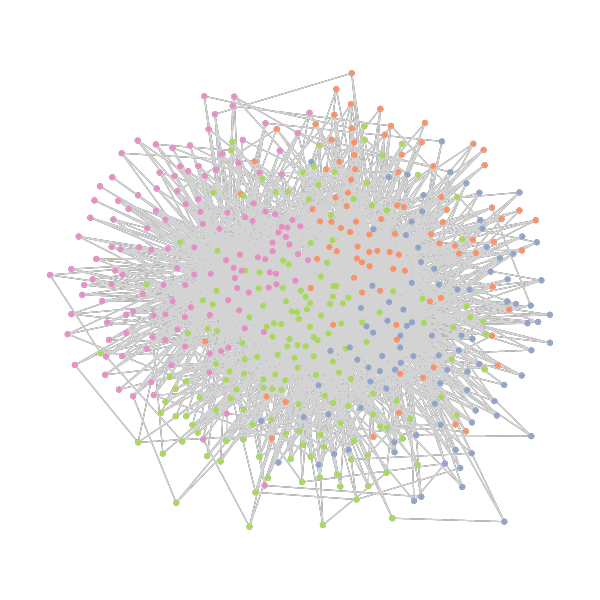}
    \end{subfigure}
\vspace{\spaceforcaption}\caption{LFR clustered graphs; $\graphmat\factnot{1}$ left, $\graphmat\factnot{2}$ middle, and
$\graphmat\factnot{3}$ right.}
\label{fig:comm_graphs}
\end{figure*}
\begin{figure*}[!h]
    \centering
    \begin{subfigure}[b]{\scalarfordc\columnwidth}
    \centering{
%
%
%
\begin{tikzpicture}

\begin{axis}[%
width=0.956\mywidth,
height=0.987\myheight,
at={(0\mywidth,0\myheight)},
scale only axis,
xmin=10,
xmax=28,
xlabel style={font=\xlabelfontsize},
xlabel={Signal to Noise Ratio (dB)},
ymin=0,
ymax=1,
ylabel style={font=\ylabelfontsize},
ylabel={NMI},
axis background/.style={fill=white},
legend columns=4,
xmajorgrids,
ymajorgrids,
grid style={dotted},ticklabel style={font=\ticklabelfontsize},
legend style={
	at={(0,1.015)}, 
	anchor=south west, legend cell align=left, align=left, draw=none
	, font=\legendfontsize}
]

\addplot [color=CGTF, line width=2.0pt,dashed
, mark=o, mark options={solid, CGTF}]
  table[row sep=crcr]{%
10	0.223021004490254\\
12	0.37830480039822\\
14	0.534318021531171\\
16	0.786142330490291\\
18	0.851047921008109\\
20	0.838367626029896\\
22	0.91351381995668\\
24	0.957919811129494\\
26	0.961592048685901\\
28	0.959935053582607\\
};
\addlegendentry{CGTF}

\addplot [color=CGTF, line width=2.0pt
, mark=diamond, mark options={solid, CGTF}]
  table[row sep=crcr]{%
10	0.214032513315889\\
12	0.391049544619434\\
14	0.577846750932454\\
16	0.682713841874355\\
18	0.885357591688773\\
20	0.871332524336306\\
22	0.978813568103431\\
24	0.976242419762025\\
26	0.937401812415059\\
28	0.97571829190109\\
};
\addlegendentry{CGTF}

\addplot [color=SNMF, line width=2.0pt, dashed
, mark=o, mark options={solid, SNMF}]
  table[row sep=crcr]{%
10	0.0755318443288618\\
12	0.0784150664661194\\
14	0.104188988524889\\
16	0.135226845587603\\
18	0.163877535829372\\
20	0.276811288908495\\
22	0.510611887217864\\
24	0.687620105007285\\
26	0.865568166330525\\
28	0.930388916213033\\
};
\addlegendentry{SNMF}

\addplot [color=SNMF, line width=2.0pt
, mark=diamond, mark options={solid, SNMF}]
  table[row sep=crcr]{%
10	0.0850346674760761\\
12	0.0728476839405747\\
14	0.115371744657463\\
16	0.125921671831722\\
18	0.152177601726959\\
20	0.30240435212669\\
22	0.536187586754623\\
24	0.695746734363534\\
26	0.815601385321583\\
28	0.928384236121808\\
};
\addlegendentry{SNMF}

\end{axis}
\end{tikzpicture}%
}
    \vspace{\spaceforcaption}\caption{$\graphmat\factnot{2}$ (dashed) and 
	$\graphmat\factnot{3}$ (solid)}
    \end{subfigure}%
    ~ 
    \begin{subfigure}[b]{\scalarfordc\columnwidth}
    \centering{
%
%
\begin{tikzpicture}

\begin{axis}[%
width=0.956\mywidth,
height=0.987\myheight,
at={(0\mywidth,0\myheight)},
scale only axis,
xmin=10,
xmax=28,
xlabel style={font=\xlabelfontsize},
xlabel={Signal to Noise Ratio (dB)},
ymin=0,
ymax=1,
ylabel style={font=\ylabelfontsize},
ylabel={NMI},
xmajorgrids,
ymajorgrids,
grid style={dotted},ticklabel style={font=\ticklabelfontsize},
axis background/.style={fill=white},
legend columns=4,
legend style={
	at={(0,1.015)}, 
	anchor=south west, legend cell align=left, align=left, 
	draw=none
	, font=\legendfontsize}]
\addplot [color=CGTF, line width=2.0pt 
, mark=o, mark options={solid, CGTF}]
  table[row sep=crcr]{%
10	0.458930301783447\\
12	0.626500342027738\\
14	0.783914167633068\\
16	0.885694050229409\\
18	0.957400193418282\\
20	0.948647296074221\\
22	0.971796072699434\\
24	1\\
26	0.995249319688767\\
28	1\\
};
\addlegendentry{CGTF}

\addplot [color=SNMF, line width=2.0pt
, mark=diamond, mark options={solid, SNMF}]
  table[row sep=crcr]{%
10	0.101520308068252\\
12	0.0933227661291904\\
14	0.0976782606774314\\
16	0.107518350753888\\
18	0.115017277078632\\
20	0.126557433985506\\
22	0.0917733350821995\\
24	0.105980944504614\\
26	0.139845953233862\\
28	0.143247292513754\\
};
\addlegendentry{SNMF}

\end{axis}
\end{tikzpicture}%
}
    \vspace{\spaceforcaption}\caption{ $\graphmat\factnot{1}$}
    \end{subfigure}
	\vspace{\spaceforcaption}\caption{Community detection performance based on 
	NMI.
	}
\label{fig:comm_graphs_nmifull}
\end{figure*}
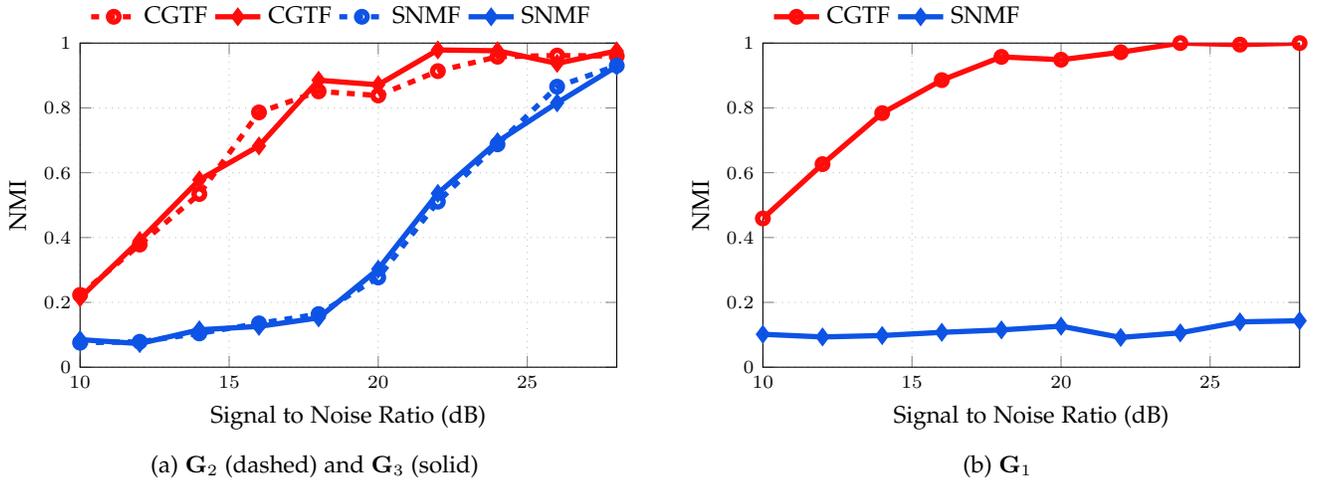

To evaluate the performance of the CGTF in detecting communities,
we employed the Lancicchinetti-Fortunato-Radicci (LFR)
benchmark~\cite{lancichinetti08bench} that generates graphs with
ground truth communities. LFR graphs capture properties of
real-world  networks such as heterogeneity in the distributions 
of node degrees and also in the community sizes.

First, we generated 3 LFR networks $\{\graphmat\factnot{\factind}\}_{\factind=1}^3$ 
with $\factorsize\factnot{1}=100$, $ 
\factorsize\factnot{2}=300$, and
$\factorsize\factnot{3}=500$ nodes, 
correspondingly comprising 
$\commnbr\factnot{1}=5$, $\commnbr\factnot{2}=3$ , $\commnbr\factnot{3}=4$ 
 communities; see Fig. \ref{fig:comm_graphs}. 
We recover the factors $\{\factmat\factnot{\factind}\}_\factind$ of 
$\{\graphmat\factnot{\factind}\}_\factind$  using SNMF, and 
construct $\datatensor$  using \eqref{eq:parafacm}. 
Next, we observe  noisy versions of the tensor data and the
corresponding graph adjacency matrices; for $\graphmat\factnot{1}$ we observe
only $10\%$ of its entries and $\ranktensor=5$.

Fig. \ref{fig:comm_graphs_nmifull}a shows the NMI performance of
CGTF and SNMF~\cite{kuang2012symmetric}, as we increase the SNR
for $\graphmat
\factnot{2}$ and $\graphmat
\factnot{3}$. The proposed approach recovers successfully robust community
assignments.

Furthermore, Fig. \ref{fig:comm_graphs_nmifull}b depicts the NMI 
performance of the algorithms with $90\%$ entries
of $\graphmat\factnot{1}$ missing. As expected, SNMF cannot recover
the community assignments of the nodes in this partially
observed $\graphmat\factnot{1}$. On the other hand, the 
novel CGTF exploits the 
tensor data, copes with missing links,
and provides reliable estimates of $\assignmentvec\factnot{1}$.

\subsection{Activities of users at different locations}
To assess the potential of our approach in providing
accurate recommendations,  we further tested a real
recommendation dataset that comprises a 
three-way tensor indicating the frequency of a user performing an activity at a 
certain location~\cite{Zheng2010CollaborativeFM}. It contains information about 164 
users, 168 locations and 5 activities. A binary tensor $\datatensor$ is 
constructed to represent the links between
users, their locations and corresponding 
activities. In other words, 
$\datatensor{(\tensind\factnot{1},\tensind\factnot{2},
	\tensind\factnot{3})}$ equals 1 if user $ \tensind\factnot{1} $ 
visited location $ \tensind\factnot{2} $ and performed activity $ 
\tensind\factnot{3} $; otherwise, it is 0. 
Additionally, similarity matrices between the users and the activities are 
provided. The similarity value between two locations is defined by the inner 
product of the corresponding feature vectors. The dataset is missing 
social network information for  28 users, and feature vectors for 32 locations. 
The parameters of CGTF 
were
	$\{\regadmm_{\cfactmat\factnot{\factind}}=100,
		\regadmm_{\pfactmat\factnot{\factind}}=100,
		\regadmm_{\pdiagvec\factnot{\factind}}=100\}_\factind, \regpar=10^{-4}$,
and for all approaches 
$\ranktensor=5$.

Table~\ref{table: NMSE} lists the NMSE for variable percentages of
missing tensor data. The CGTF model  
exploits judiciously the structure of the 
available graph information, which enables our efficient ADMM solver
to outperform 
competing alternatives, and lead to improved recommendations. 

\begin{table}[!h]
\centering
\begin{tabular}{||c | c c c c||} 
\hline
Missing &  NTF & PARAFAC & CMTF & CGTF \\ 
\hline\hline
40\% &  0.995 & 1.016& 0.98  & $\bm{ 0.46}$   \\ 
\hline
50\% & 0.99 & 0.96& 0.99& $\bm{ 0.68}$   \\
\hline
\end{tabular}
\caption[]{NMSE for different ratios of missing data.}
\label{table: NMSE}
\end{table}

In order to assess the recommendation quality of the proposed approach, 
we changed the thereshold for detecting an activity (edge) 
on the tensor (graphs). Per threshold value, we then obtained  the probabilities
of detection and false alarm.

\begin{figure*}[!h]
    \centering
    \begin{subfigure}[b]{\scalarfordc\columnwidth}
    \centering{\input{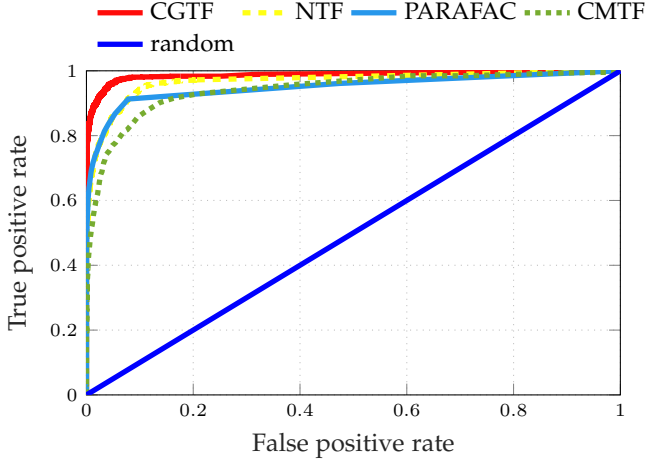}}
    \end{subfigure}%
    ~ 
    \begin{subfigure}[b]{\scalarfordc\columnwidth}
    \centering{\input{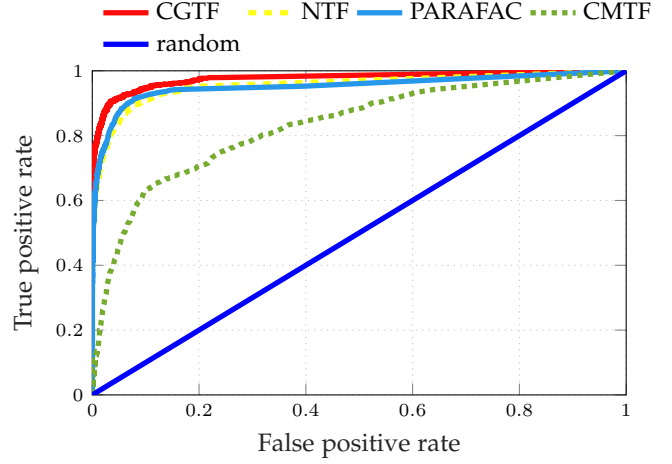}}
    \end{subfigure}
	\vspace{\spaceforcaption}\caption{ROC for 40\%  (left); and 50\% (right)
	tensor missing entries.}
\label{fig:tensorrocr1}
\end{figure*}

Fig. \ref{fig:tensorrocr1} depicts the  receiver operating characteristic 
(ROC) for  the tensor entries, and as expected
the novel CGTF outperforms the alternative.
Moreover, Fig. \ref{fig:graphroc} shows the ROC for
discovering concealed links in the user-graph
with only $10\%$ of observed graph entries
when the factors are initialized either using the SNMF
or randomly. In both cases, CGTF performs successful
edge identification and yields accurate
link predictions. The performance gap
among CGTF and CMTF, becomes more pronounced
when the factors are initialized randomly, which suggests that
initialization is crucial in achieving a good stationary point.

\begin{figure*}[!h]
    \centering
    \begin{subfigure}[b]{\scalarfordc\columnwidth}
    \centering{\input{figs/graph1roc40X90G190G2init.tex}}
    \end{subfigure}%
    ~ 
    \begin{subfigure}[b]{\scalarfordc\columnwidth}
    \centering{\input{figs/graph1roc40X90G190G2randinit.tex}}
    \end{subfigure}
	\vspace{\spaceforcaption}\caption{ROC curve  for
	$\graphmat\factnot{1}$ using
	the SNMF for initialization of $\factmat\factnot{1}$(left),
	random initialization
	(right)
	with $40\%$ misses in $\datatensor$ 
	and $90\%$ misses in $\graphmat\factnot{1}$
	and $\graphmat\factnot{2}$.}
\label{fig:graphroc}
\end{figure*}

\subsubsection{\change{{Community detection}}}
Furthermore, CD is pursued for the user and 
location graphs with 70\% entries missing 
in the tensor and no misses in the graphs.
\change{We compare our CD performance against the following baselines: Potts~\cite{ronhovde2010local}, NewmanF~\cite{danon2006effect}, SP~\cite{hespanha2004efficient}, AFG~\cite{arenas2008analysis} and SNMF~\cite{kuang2012symmetric}\footnote{\change{We use the Matlab implementations provided by the authors.}}.} 
In lieu of ground-truth communities, we evaluate the CD performance by the maximum conductance-coverage curve. This curve is plotted by varying $\covscalar$ from 0 to 1 (cf. \eqref{eq:cov})  and reporting the corresponding coverage value on the x-axis and maximum conductunce the y-axis.
Low values of conductance for large values of coverage correspond to more cohesive communities. Hence, a smaller area under curve (AUC) implies better performance; see Sec. \ref{sec:com_det_eval}.
\change{Fig. \ref{fig:comm_graphs_cover} reports the coverage scores relative to the maximum conductance ($\covscalar$) for the users graph (left) and the locations graph (right).
The proposed CGTF achieves higher coverage scores for smaller conductance and outperforms competing approaches. CGTF achieves the smallest AUC value in the user graph and one of the smallest in the location graph. Hence, the factors obtained by our coupled approach indeed improve CD performance.}

\begin{figure*}[h!]
    \centering
    \begin{subfigure}[b]{\scalarfordc\columnwidth}
    \centering{
%
%
\definecolor{mycolor1}{rgb}{0.00000,0.44700,0.74100}%
\definecolor{mycolor2}{rgb}{0.85000,0.32500,0.09800}%
\definecolor{mycolor3}{rgb}{0.92900,0.69400,0.12500}%
\definecolor{mycolor4}{rgb}{0.49400,0.18400,0.55600}%
\definecolor{mycolor5}{rgb}{0.46600,0.67400,0.18800}%
\definecolor{mycolor6}{rgb}{0.30100,0.74500,0.93300}%
\begin{tikzpicture}

\begin{axis}[%
width=0.956\mywidth,
height=0.987\myheight,
at={(0\mywidth,0\myheight)},
scale only axis,
xmin=0,
xmax=1,
xlabel style={font=\color{white!15!black}},
xlabel={Coverage},
ymin=0,
ymax=1,
ylabel style={font=\color{white!15!black}},
ylabel={Maximum conductunce},
grid style={dotted},ticklabel style={font=\ticklabelfontsize},
legend columns=4,
legend style={
	at={(0,1.015)}, 
	anchor=south west, legend cell align=left, align=left, draw=none
	, font=\legendfontsize}
]
\addplot [color=CGTF, mark=*, line width=2.0pt]
  table[row sep=crcr]{%
0	0\\
0	0.1\\
0.24	0.2\\
0.44	0.3\\
0.684146341463415	0.4\\
0.684146341463415	0.5\\
0.794634146341463	0.6\\
0.794634146341463	0.7\\
1	0.8\\
1	0.9\\
1	1\\
};
\addlegendentry{CGTF}

\addplot [color=SNMF,dashdotted, mark=triangle*, line width=2.0pt]
  table[row sep=crcr]{%
0	0\\
0	0.1\\
0	0.2\\
0	0.3\\
0.26219512195122	0.4\\
0.50609756097561	0.5\\
0.50609756097561	0.6\\
0.896341463414634	0.7\\
1	0.8\\
1	0.9\\
1	1\\
};
\addlegendentry{NMF}

\addplot [color=AFG, mark=star, line width=2.0pt]
  table[row sep=crcr]{%
0	0\\
0.0182926829268293	0.1\\
0.0182926829268293	0.2\\
0.315	0.3\\
0.315 0.4\\
0.615	0.5\\
0.615	0.6\\
1	0.7\\
1	0.8\\
1	0.9\\
1	1\\
};
\addlegendentry{AFG}

\addplot [color=NewmanF, mark=otimes*, line width=2.0pt]
  table[row sep=crcr]{%
0	0\\
0.0182926829268293	0.1\\
0.0182926829268293	0.2\\
0.325	0.3\\
0.325	0.4\\
0.325	0.5\\
0.625	0.6\\
0.625	0.7\\
1	0.8\\
1	0.9\\
1	1\\
};
\addlegendentry{NewmanF}

\addplot [color=SP, dotted, mark=square*, line width=2.0pt]
  table[row sep=crcr]{%
0	0\\
0.170731707317073	0.1\\
0.170731707317073	0.2\\
0.39390243902439	0.3\\
0.641463414634146	0.4\\
0.641463414634146	0.5\\
0.641463414634146	0.6\\
0.841463	0.7\\
0.841463	0.8\\
1	0.9\\
1	1\\
};
\addlegendentry{SP}

\addplot [color=Pott, dashed, mark=diamond*,line width=2.0pt]
  table[row sep=crcr]{%
0	0\\
0.0182926829268293	0.1\\
0.0182926829268293	0.2\\
0.0182926829268293	0.3\\
0.0182926829268293	0.4\\
0.0182926829268293	0.5\\
1	0.6\\
1	0.7\\
1	0.8\\
1	0.9\\
1	1\\
};
\addlegendentry{Potts}

\end{axis}
\end{tikzpicture}%
}
    \end{subfigure}%
    ~ 
    \begin{subfigure}[b]{\scalarfordc\columnwidth}
    \centering{
%
%
\definecolor{mycolor1}{rgb}{0.00000,0.44700,0.74100}%
\definecolor{mycolor2}{rgb}{0.85000,0.32500,0.09800}%
\definecolor{mycolor3}{rgb}{0.92900,0.69400,0.12500}%
\definecolor{mycolor4}{rgb}{0.49400,0.18400,0.55600}%
\definecolor{mycolor5}{rgb}{0.46600,0.67400,0.18800}%
\definecolor{mycolor6}{rgb}{0.30100,0.74500,0.93300}%
\begin{tikzpicture}

\begin{axis}[%
width=0.956\mywidth,
height=0.987\myheight,
at={(0\mywidth,0\myheight)},
scale only axis,
xmin=0,
xmax=1,
xlabel style={font=\color{white!15!black}},
xlabel={Coverage},
ymin=0,
ymax=1,
ylabel style={font=\color{white!15!black}},
ylabel={Maximum conductunce},
grid style={dotted},ticklabel style={font=\ticklabelfontsize},
legend columns=4,
legend style={
	at={(0,1.015)}, 
	anchor=south west, legend cell align=left, align=left, draw=none
	, font=\legendfontsize}
]
\addplot [color=CGTF, mark=*,line width=2.0pt]
  table[row sep=crcr]{%
0	0\\
0	0.1\\
0	0.2\\
0	0.3\\
0.4	0.4\\
0.62380952381	0.5\\
0.75238095238095	0.6\\
0.8047619047619	0.7\\
0.964047619047619	0.8\\
0.976666666666667	0.9\\
1	1\\
};
\addlegendentry{CGTF}

\addplot [color=SNMF,dashdotted, mark=triangle*, line width=2.0pt]
  table[row sep=crcr]{%
0	0\\
0	0.1\\
0	0.2\\
0	0.3\\
0	0.4\\
0.19047619047619	0.5\\
0.333333333333333	0.6\\
0.488095238095238	0.7\\
0.708333333333333	0.8\\
0.994047619047619	0.9\\
0.994047619047619	1\\
};
\addlegendentry{SNMF}

\addplot [color=AFG, mark=star, line width=2.0pt]
  table[row sep=crcr]{%
0	0\\
0	0.1\\
0	0.2\\
0	0.3\\
0	0.4\\
0.785714285714286	0.5\\
0.964285714285714	0.6\\
0.964285714285714	0.7\\
0.964285714285714	0.8\\
0.964285714285714	0.9\\
0.964285714285714	1\\
};
\addlegendentry{AFG}

\addplot [color=NewmanF, mark=otimes*,  line width=2.0pt]
  table[row sep=crcr]{%
0	0\\
0	0.1\\
0	0.2\\
0	0.3\\
0.333333333333333	0.4\\
0.732142857142857	0.5\\
0.916666666666667	0.6\\
0.916666666666667	0.7\\
0.916666666666667	0.8\\
0.964285714285714	0.9\\
0.964285714285714	1\\
};
\addlegendentry{NewmanF}

\addplot [color=SP, dotted, mark=square*, line width=2.0pt]
  table[row sep=crcr]{%
0	0\\
0	0.1\\
0	0.2\\
0	0.3\\
0	0.4\\
0	0.5\\
0.55952380952381	0.6\\
0.55952380952381	0.7\\
0.696428571428571	0.8\\
0.827380952380952	0.9\\
0.827380952380952	1\\
};
\addlegendentry{SP}

\addplot [color=Pott, dashed, mark=diamond*,line width=2.0pt]
  table[row sep=crcr]{%
0	0\\
0	0.1\\
0	0.2\\
0	0.3\\
0	0.4\\
0	0.5\\
0.952380952380952	0.6\\
0.952380952380952	0.7\\
0.952380952380952	0.8\\
0.952380952380952	0.9\\
0.952380952380952	1\\
};
\addlegendentry{Potts}

\end{axis}
\end{tikzpicture}%
}
    \end{subfigure}
	\vspace{\spaceforcaption}\caption{\change{Community detection performance based on coverage 
	for the user graph (left); and the location graph (right).}}
\label{fig:comm_graphs_cover}
\end{figure*}

\subsection{Posts of users in a social network}
We also tested the performance of  CGTF on the Digg dataset. 
Digg is a social network that allows users to submit, Digg, and comment 
on news stories. In~\cite{lin2009metafac}, the data was collected from a 
large subset of users and stories. The dataset includes stories, and users 
along with their time-stamped actions with respect to stories, as well as the social
network of users. In addition, a set of keywords is assigned to each story.

After discretizing the time into $20$ time intervals over $3$ days,
we construct a tensor comprising the number of comments that user $i$ wrote on story
$j$ during the $k$-th time interval stored in the $(i, j, k)$ item. Also, a
story-story graph is constructed where any two stories are connected only if 
they share more than two keywords. The original tensor containing all users and stories
includes a large number of inactive users and unpopular stories. In order to assess 
performance of the proposed method, the data is subsampled so that 
the $175$ most active users and the $800$ most popular stories are kept. Hence, the size
of the tensor in this experiment is $\factorsize\factnot{1}=175$ users,
$\factorsize\factnot{2}=800$ stories and $\factorsize\factnot{2}=20$ time intervals. 
In addition, the side  information comprises two graphs that represent the users' social 
network and the similarities of the stories.

The tensor and the two graphs are fused jointly as in~\eqref{eq:cgtf} with $\ranktensor=10$.
Then, the proposed ADMM-based algorithm is employed to obtain the latent factors of the CGTF model.
As there is no graph on the third mode (time intervals),
the term $\|\graphmat \factnot{3} 
-\factmat\factnot{3}\diag{(\diagvec\factnot{3})}\factmat\factnot{3}^
 \transpose\|^2_F$ is not included in~\eqref{eq:cgtf}. 
 We assume that $ 40\%$ of the  tensor entries, as well as  $ 30\%$ of the links
 in the user-user and story-story graphs are missing.
 
\begin{figure}[!h]
\centering{\input{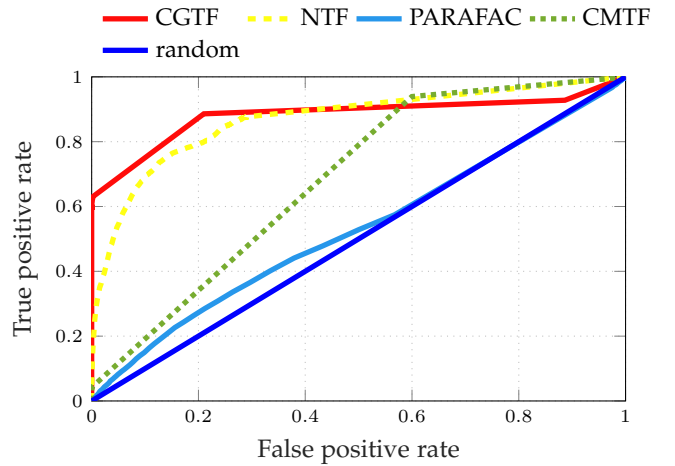}  
}
	\vspace{\spaceforcaption}\caption{ROC for 40\%  
	tensor missing entries.}
\label{fig:tensorroc}
\end{figure}
\begin{figure*}[!h]
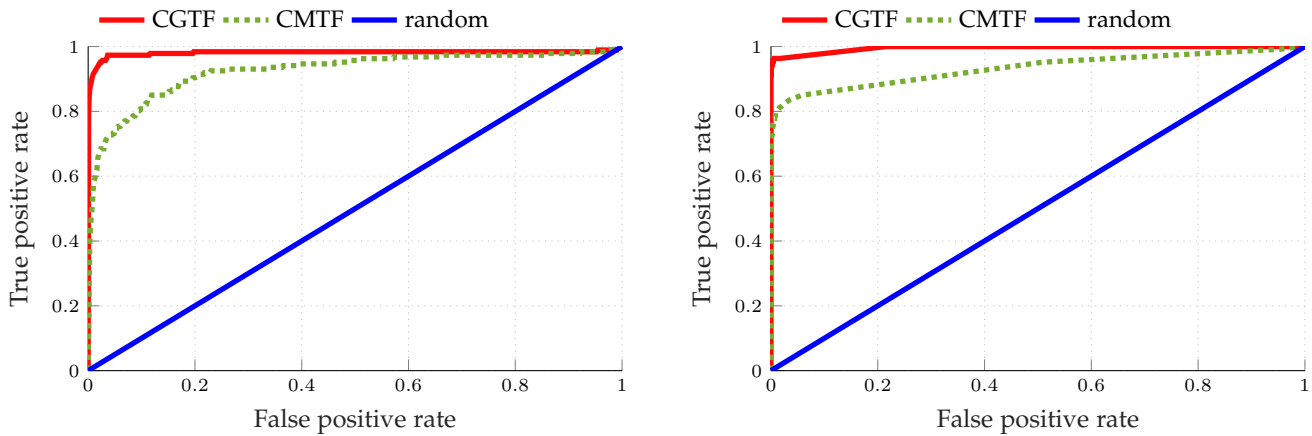

    \centering
    \begin{subfigure}{\scalarfordc\columnwidth}
\centering{\input{figs/graph1r2roc40X30G130G2randinit.tex}}    
    \end{subfigure}%
    ~ 
    \begin{subfigure}{\scalarfordc\columnwidth}
  \centering{\input{figs/graph2r2roc40X30G130G2randinit.tex}}
    \end{subfigure}%
	\vspace{\spaceforcaption}\caption{ROC for the prediction in the 
	users' social network $\graphmat\factnot{1}$
	(left); and the story graph adjacency 
	$\graphmat\factnot{2}$ (right).}
\label{fig:graphs_roc}
\end{figure*}

In Figs.~\ref{fig:tensorroc}, and~\ref{fig:graphs_roc},
the ROC is presented for the tensor and the graphs. The proposed
approach outperforms competing approaches in completing the missing tensor entries as well as predicting the missing links in the graph, 
and leads to accurate recommendations for previously unseen data.

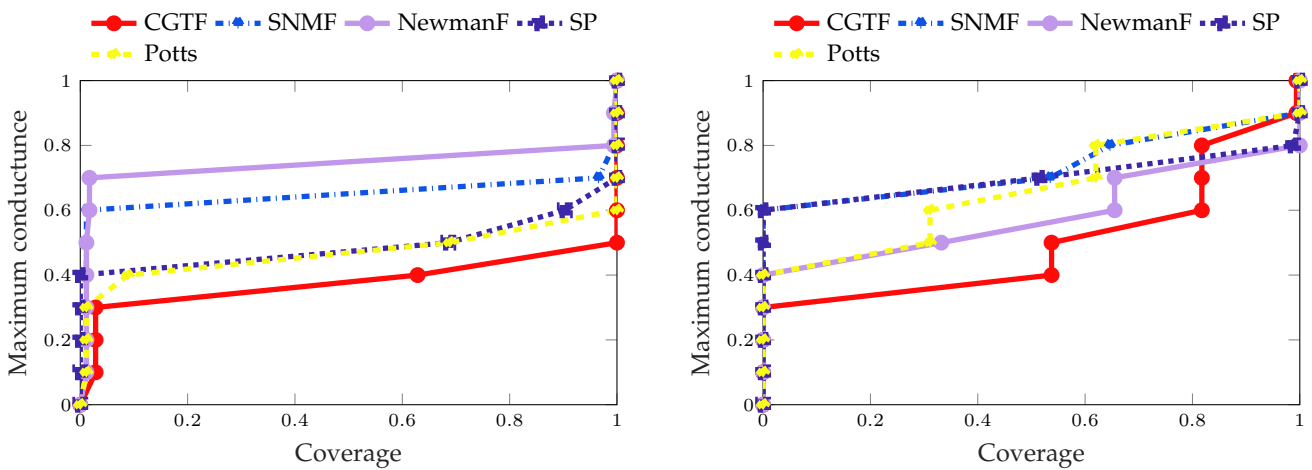
\begin{figure*}[!h]
    \centering
    \begin{subfigure}[b]{\scalarfordc\columnwidth}
    \centering{
%
%
\definecolor{mycolor1}{rgb}{0.00000,0.44700,0.74100}%
\definecolor{mycolor2}{rgb}{0.85000,0.32500,0.09800}%
\definecolor{mycolor3}{rgb}{0.92900,0.69400,0.12500}%
\definecolor{mycolor4}{rgb}{0.49400,0.18400,0.55600}%
\definecolor{mycolor5}{rgb}{0.46600,0.67400,0.18800}%
\begin{tikzpicture}

\begin{axis}[%
width=0.956\mywidth,
height=0.987\myheight,
at={(0\mywidth,0\myheight)},
scale only axis,
xmin=0,
xmax=1,
xlabel style={font=\color{white!15!black}},
xlabel={Coverage},
ymin=0,
ymax=1,
ylabel style={font=\color{white!15!black}},
ylabel={Maximum conductunce},
grid style={dotted},ticklabel style={font=\ticklabelfontsize},
legend columns=4,
legend style={
	at={(0,1.015)}, 
	anchor=south west, legend cell align=left, align=left, draw=none
	, font=\legendfontsize}
]
\addplot [color=CGTF, mark=*, line width=2.0pt]
  table[row sep=crcr]{%
0	0\\
0.0285714285714286	0.1\\
0.0285714285714286	0.2\\
0.0285714285714286	0.3\\
0.62857142857143	0.4\\
1	0.5\\
1	0.6\\
1	0.7\\
1	0.8\\
1	0.9\\
1	1\\
};
\addlegendentry{CGTF}

\addplot[color=SNMF,dashdotted, mark=triangle*,  line width=2.0pt]
  table[row sep=crcr]{%
0	0\\
0.0114285714285714	0.1\\
0.0114285714285714	0.2\\
0.0114285714285714	0.3\\
0.0114285714285714	0.4\\
0.0114285714285714	0.5\\
0.0114285714285714	0.6\\
0.965714285714286	0.7\\
0.994285714285714	0.8\\
1	0.9\\
1	1\\
};
\addlegendentry{SNMF}

\addplot  [color=NewmanF, mark=otimes*, line width=2.0pt]
  table[row sep=crcr]{%
0	0\\
0.0114285714285714	0.1\\
0.0114285714285714	0.2\\
0.0114285714285714	0.3\\
0.0114285714285714	0.4\\
0.0114285714285714	0.5\\
0.0171428571428571	0.6\\
0.0171428571428571	0.7\\
0.994285714285714	0.8\\
0.994285714285714	0.9\\
1	1\\
};
\addlegendentry{NewmanF}

\addplot [color=SP, dotted, mark=square*, line width=2.0pt]
  table[row sep=crcr]{%
0	0\\
0	0.1\\
0	0.2\\
0	0.3\\
0	0.4\\
0.685714285714286	0.5\\
0.902857142857143	0.6\\
1	0.7\\
1	0.8\\
1	0.9\\
1	1\\
};
\addlegendentry{SP}

\addplot [color=Pott, dashed, mark=diamond*, line width=2.0pt]
  table[row sep=crcr]{%
0	0\\
0.0114285714285714	0.1\\
0.0114285714285714	0.2\\
0.0114285714285714	0.3\\
0.088571428571429	0.4\\
0.688571428571429	0.5\\
1	0.6\\
1	0.7\\
1	0.8\\
1	0.9\\
1	1\\
};
\addlegendentry{Potts}

\end{axis}
\end{tikzpicture}%
}
    \end{subfigure}%
    ~ 
    \begin{subfigure}[b]{\scalarfordc\columnwidth}
    \centering{
%
%
%
\begin{tikzpicture}

\begin{axis}[%
width=0.956\mywidth,
height=0.987\myheight,
at={(0\mywidth,0\myheight)},
scale only axis,
xmin=0,
xmax=1,
xlabel style={font=\color{white!15!black}},
xlabel={Coverage},
ymin=0,
ymax=1,
ylabel style={font=\color{white!15!black}},
ylabel={Maximum conductunce},
grid style={dotted},ticklabel style={font=\ticklabelfontsize},
legend columns=4,
legend style={
	at={(0,1.015)}, 
	anchor=south west, legend cell align=left, align=left, draw=none
	, font=\legendfontsize}
]
\addplot [color=CGTF,mark=*, line width=2.0pt]
  table[row sep=crcr]{%
0	0\\
0	0.1\\
0	0.2\\
0	0.3\\
0.5375	0.4\\
0.5375	0.5\\
0.8175	0.6\\
0.8175	0.7\\
0.8175	0.8\\
0.99375	0.9\\
0.99375	1\\
};
\addlegendentry{CGTF}

\addplot [color=SNMF,dashdotted, mark=triangle*, line width=2.0pt]
  table[row sep=crcr]{%
0	0\\
0	0.1\\
0	0.2\\
0	0.3\\
0	0.4\\
0	0.5\\
0	0.6\\
0.535	0.7\\
0.64625	0.8\\
1	0.9\\
1	1\\
};
\addlegendentry{SNMF}

\addplot [color=NewmanF, mark=otimes*, line width=2.0pt]
  table[row sep=crcr]{%
0	0\\
0	0.1\\
0	0.2\\
0	0.3\\
0	0.4\\
0.332	0.5\\
0.655	0.6\\
0.655	0.7\\
1	0.8\\
1	0.9\\
1	1\\
};
\addlegendentry{NewmanF}

\addplot [color=SP, dotted, mark=square*, line width=2.0pt]
  table[row sep=crcr]{%
0	0\\
0	0.1\\
0	0.2\\
0	0.3\\
0	0.4\\
0	0.5\\
0	0.6\\
0.515	0.7\\
0.9875	0.8\\
1	0.9\\
1	1\\
};
\addlegendentry{SP}

\addplot [color=Pott, dashed, mark=diamond*, line width=2.0pt]
  table[row sep=crcr]{%
0	0\\
0	0.1\\
0	0.2\\
0	0.3\\
0	0.4\\
0.311	0.5\\
0.311	0.6\\
0.62	0.7\\
0.62	0.8\\
1	0.9\\
1	1\\
};
\addlegendentry{Potts}

\end{axis}
\end{tikzpicture}%
}
    \end{subfigure}
	\vspace{\spaceforcaption}\caption{\change{Community detection performance based on coverage 
	for the user graph (left); and the story graph (right).}}
\label{fig:comm_graphs_cover2}
\end{figure*}

\subsubsection{\change{{Community detection under missing links}}}\change{
In this experiment we assume that $ 40\%$ of the  tensor entries and  $50\%$ of  the graph links are missing. The goal here is to examine whether CGTF  recovers the communities in the graphs even with hidden graph links. Fig. \ref{fig:comm_graphs_cover2} reports the coverage scores relative to the maximum conductance for the users graph (left) and the stories graph (right).\footnote{AFG did not provide meaningful results and was not included.} Competing approaches that only utilize the partially observed graphs can not recover crisp graph communities. On the other hand,  our novel CGTF utilizes judiciously the partially observed graphs and tensors and reports superior performance. The advantage of the proposed framework in community detection is more evident in this experiment (compare Fig. \ref{fig:comm_graphs_cover2} and Fig. \ref{fig:comm_graphs_cover}).}

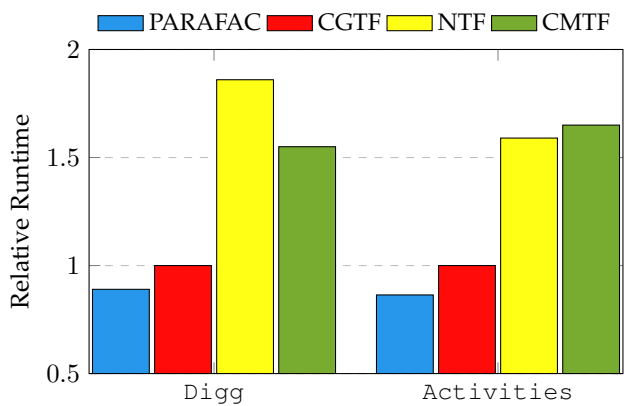
\begin{figure}[!h]
\centering{
%

%
\begin{tikzpicture}

\begin{axis}[%
width=0.951\mywidth,
height=0.987\myheight,
at={(0\mywidth,0\myheight)},
scale only axis,
bar shift auto,
log origin=infty,
xmin=-1.2,
xmax=8.2,
xtick={1,6},
xticklabels={{\texttt{Digg},\texttt{Activities}}
},
ymin=0.5,
ymax=2,
yminorticks=true,
ylabel style={font=\color{white!15!black}},
ylabel={Relative Runtime},
ylabel near ticks,
axis background/.style={fill=white},
ymajorgrids,
grid style={dashed},
legend columns=4,
legend style={
	at={(0,1.015)}, 
	anchor=south west, legend cell align=left, align=left, draw=none
	, font=\legendfontsize}
]

\addplot[ybar, bar width=1, fill=PARAFAC, draw=black, area legend] table[row sep=crcr] {%
1	0.89\\
6	0.864\\
};
\addlegendentry{PARAFAC}

\addplot[ybar, bar width=1, fill=CGTF, draw=black, area legend] table[row sep=crcr] {%
	1	1\\
    6	1\\
};
\addlegendentry{CGTF}

\addplot[ybar, bar width=1, fill=NTF, draw=black, area legend] table[row sep=crcr] {%
	1	1.86\\
	6	1.59\\
};
\addlegendentry{NTF}

\addplot[ybar, bar width=1, fill=CMTF, draw=black, area legend] table[row sep=crcr] {%
	1	1.55\\
	6	1.65\\
};
\addlegendentry{CMTF}

%
%
%

\end{axis}
\end{tikzpicture}%
}
	\vspace{\spaceforcaption}\caption{\change{Runtime comparisons relative to CGTF.}}
\label{fig:runtime}
\end{figure}

\subsection{\change{Runtime  comparisons}}
\change{The scalabilty of CGTF is reflected on the relative runtime comparisons
listed in Fig. \ref{fig:runtime}, for recovering the tensor entries for the Activities and Digg datasets  in Fig.~\ref{fig:tensorrocr1} and Fig.~\ref{fig:tensorroc} respectively.  
All experiments were run on a machine with i7-4790 @3.60 Ghz CPU, and 32GB of RAM. We used the Matlab implementations provided by the authors of the compared algorithms. The bars in Fig. \ref{fig:runtime} indicate the runtime of the algorithms relative to CGTF's runtime. 
Evidently, our efficient yet effective CGTF implementation is almost as fast as the PARAFAC, while achieving superior tensor imputation performance (see Figs.~\ref{fig:tensorrocr1},\ref{fig:tensorroc}).}

\section{Conclusions and future work}
This paper investigates the inference of unavailable
entries in tensors and graphs based on a novel CGTF model.
An efficient algorithm is developed to identify the factor matrices 
and recover the missing entries.  The ADMM solver features closed-form updates and is amenable 
to parallel and accelerated implementation. In addition, the proposed method 
can overcome the so-called {\em cold-start} problem, where the tensor
has missing slabs or the similarity 
matrices are not complete. The novel algorithm makes accurate prediction of the 
missing values and can be used in many real world settings, especially in 
recommender systems. 
A novel direction is further explored by employing the interpretable factors
of CGTF to detect communities of nodes in the
graphs having the tensor as side information.
Through numerical tests with synthetic
as well as real-data, the novel algorithm was observed to
perform markedly better than existing alternatives and further yield accurate
recommendations, as well as effective identification of communities.

\change{
Our future research agenda will focus in two direction.  Todays era of data deluge has grown the interest for robust methods that can handle anomalies in collections of high-dimensional data. Towards this end, we aim at a robust CGTF to handle anomalies in the tensor and graph data.  Furthermore, in many scenarios prior information on the tensor and graph data can be accounted for to improve imputation performance. CGTF may incorporate such knowledge by introducing a probabilistic prior for certain graphs e.g. stochastic block models~\cite{airoldi2008mixed}}.

\section{Proof of Proposition \ref{prop:conv}}
\label{proof:prop}
In what follows, we omit the terms $\datatensormiss,\{\graphmatmiss\factnot{\factind}
\}_{\factind=1}^3$, although the proof can be easily modified to accommodate misses in the graphs
and in the tensor.
First, we claim 
\begin{align}
\label{eq:conv}
\collectedvar\admmiternot{\admmiter+1}-
\collectedvar\admmiternot\admmiter\rightarrow\mathbf{0}\nonumber,\\
\collecteddualvar\admmiternot{\admmiter+1}-
\collecteddualvar\admmiternot\admmiter\rightarrow\mathbf{0}.
\end{align} 
Observe 
that the Lagragian $\lagragian(\collectedvar,\collecteddualvar)$ is 
bounded from below which follows because
\begin{align*}
&\lagragian\big(\collectedvar,\collecteddualvar\big) 
:=\fitfun\big(\collectedvar\big)+
\sum_{f=1}^{3}\big\{ 
\tfrac{\regadmm_{\cfactmat\factnot{\factind}}}{2}\|\factmat\factnot{\factind}
- \cfactmat\factnot{\factind}+\tfrac{\dualadmmmat_{\cfactmat\factnot{\factind}}}
{\regadmm_{\cfactmat\factnot{\factind}}}\|_F^2
\end{align*}
\begin{align*}
-&\tfrac{1}{2\regadmm_{\cfactmat\factnot{\factind}}}\|\dualadmmmat_
{\cfactmat\factnot{\factind}}\|_F^2+
\tfrac{\regadmm_{\pfactmat\factnot{\factind}}}{2}\|\factmat\factnot{\factind}
- \pfactmat\factnot{\factind}+\tfrac{\dualadmmmat_{\pfactmat\factnot{\factind}}}
{\regadmm_{\pfactmat\factnot{\factind}}}\|_F^2
\\-&\tfrac{1}{2\regadmm_{\pfactmat\factnot{\factind}}}\|\dualadmmmat_
{\pfactmat\factnot{\factind}}\|_F^2
+\tfrac{\regadmm_{\pdiagvec\factnot{\factind}}}{2}\|\diagvec\factnot{\factind}
- \pdiagvec\factnot{\factind}+\tfrac{\dualadmmvec_{\pdiagvec\factnot{\factind}}}
{\regadmm_{\pdiagvec\factnot{\factind}}}\|_2^2
-\tfrac{1}{2\regadmm_{\pdiagvec\factnot{\factind}}}\|\dualadmmvec_{\pdiagvec\factnot{\factind}}\|_2^2
\big\}\nonumber
\end{align*}
and $\collecteddualvar$ is bounded. Owing to the appropriate reformulation \eqref{eq:cgtfeq},
$\lagragian(\cdot)$ is strongly convex w.r.t. each matrix variable 
$\mathbf{V}\in{\{\factmat\factnot{\factind},\cfactmat\factnot{\factind},
\pfactmat\factnot{\factind},\diagvec\factnot{\factind},\pdiagvec\factnot{\factind}
\}_{\factind=1}^3}$ separately. 
As a result,  it holds for $\mathbf{V}$ that
\begin{align}
\lagragian(\mathbf{V}+\delta\mathbf{V})
-\lagragian(\mathbf{V})\ge
\partial_{\mathbf{V}}\lagragian(\mathbf{V})^\transpose
\delta\mathbf{V}+\rho\|\delta\mathbf{V}\|_F^2
\label{eq:strconv}
\end{align}
where $\rho$ is a properly selected parameter, while
the variables except $\mathbf{V}$ are kept the same. Moreover,
if $ \mathbf{V}^\ast:=\arg\min_{\mathbf{V}}\lagragian(\mathbf{V})$ it 
follows that $\partial_{\mathbf{V}}\lagragian(\mathbf{V}^\ast)^\transpose
\delta\mathbf{V}\ge0$. Hence, for $\delta\mathbf{V}=\mathbf{V}\admmiternot{\admmiter}
-\mathbf{V}\admmiternot{\admmiter+1}$ and since $\mathbf{V}\admmiternot{\admmiter+1}:=\arg\max_{\mathbf{V}}
\lagragian(\mathbf{V})$ at the $\admmiter$-th iteration, it follows from
\eqref{eq:strconv} that
\begin{align}
\lagragian(\mathbf{V}\admmiternot{\admmiter})
-\lagragian(\mathbf{V}\admmiternot{\admmiter+1})\ge
\rho\|\mathbf{V}\admmiternot{\admmiter}-\mathbf{V}\admmiternot{\admmiter+1}\|_F^2
\label{eq:helpeq}
\end{align}
Specifying \eqref{eq:helpeq} to each variable in $\collectedvar$, yields for $\factind=1,2,3$
\begin{subequations}
\begin{align}
\lagragian(\factmat\factnot{\factind}\admmiternot{\admmiter})
-\lagragian(\factmat\factnot{\factind}\admmiternot{\admmiter+1})&\ge
\frac{\regadmm_{\cfactmat\factnot{\factind}}+\regadmm_{\pfactmat
\factnot{\factind}}}{2}\|\factmat\factnot{\factind}\admmiternot
{\admmiter}-\factmat\factnot{\factind}\admmiternot{\admmiter+1}\|_F^2\\
\lagragian(\cfactmat\factnot{\factind}\admmiternot{\admmiter})
-\lagragian(\cfactmat\factnot{\factind}\admmiternot{\admmiter+1})&\ge
\frac{\regadmm_{\cfactmat\factnot{\factind}}}{2}\|\cfactmat\factnot{\factind}\admmiternot
{\admmiter}-\cfactmat\factnot{\factind}\admmiternot{\admmiter+1}\|_F^2\\
\lagragian(\pfactmat\factnot{\factind}\admmiternot{\admmiter})
-\lagragian(\pfactmat\factnot{\factind}\admmiternot{\admmiter+1})&\ge
\frac{\regadmm_{\pfactmat
\factnot{\factind}}}{2}\|\pfactmat\factnot{\factind}\admmiternot
{\admmiter}-\pfactmat\factnot{\factind}\admmiternot{\admmiter+1}\|_F^2\\
\lagragian(\diagvec\factnot{\factind}\admmiternot{\admmiter})
-\lagragian(\diagvec\factnot{\factind}\admmiternot{\admmiter+1})&\ge
\frac{\regadmm_{\pdiagvec
\factnot{\factind}}}{2}\|\diagvec\factnot{\factind}\admmiternot
{\admmiter}-\diagvec\factnot{\factind}\admmiternot{\admmiter+1}\|_F^2\\
\lagragian(\pdiagvec\factnot{\factind}\admmiternot{\admmiter})
-\lagragian(\pdiagvec\factnot{\factind}\admmiternot{\admmiter+1})&\ge
\frac{\regadmm_{\pdiagvec
\factnot{\factind}}}{2}\|\pdiagvec\factnot{\factind}\admmiternot
{\admmiter}-\pdiagvec\factnot{\factind}\admmiternot{\admmiter+1}\|_F^2
\end{align}
\end{subequations}
It follows then for $R:=\min{\{\regadmm_{\cfactmat
\factnot{\factind}},\regadmm_{\pfactmat
\factnot{\factind}},\regadmm_{\pdiagvec
\factnot{\factind}}\}_n}$ that
\begin{align}
\lagragian(\collectedvar\admmiternot{\admmiter},\collecteddualvar\admmiternot{\admmiter})-
\lagragian(\collectedvar\admmiternot{\admmiter+1},\collecteddualvar\admmiternot{\admmiter})\ge
R\|\collectedvar\admmiternot{\admmiter}-\collectedvar\admmiternot{\admmiter+1}\|_F^2.
\label{eq:primalbound}
\end{align}
On the other hand, it holds for the dual variables that 
\begin{subequations}
\begin{align}
\lagragian(\dualadmmmat_{\cfactmat\factnot{\factind}}\admmiternot{\admmiter})
-\lagragian(\dualadmmmat_{\cfactmat\factnot{\factind}}\admmiternot{\admmiter+1})&=
\Tr(\dualadmmmat_{\cfactmat\factnot{\factind}}\admmiternot{\admmiter}-
\dualadmmmat_{\cfactmat\factnot{\factind}}\admmiternot{\admmiter+1})^\transpose(
\cfactmat\factnot{\factind}\admmiternot{\admmiter}-
\cfactmat\factnot{\factind}\admmiternot{\admmiter+1})\nonumber\\
&=-\frac{1}{\regadmm_{\cfactmat\factnot{\factind}}}\|\dualadmmmat_{
\cfactmat\factnot{\factind}}\admmiternot{\admmiter}-
\dualadmmmat_{\cfactmat\factnot{\factind}}\admmiternot{\admmiter+1}\|_F^2
\end{align}
where the last equality follows from \eqref{eq:lagrangupd}, and similarly
\begin{align}
\lagragian(\dualadmmmat_{\pfactmat\factnot{\factind}}\admmiternot{\admmiter})
-\lagragian(\dualadmmmat_{\pfactmat\factnot{\factind}}\admmiternot{\admmiter+1})
&=-\frac{1}{\regadmm_{\pfactmat\factnot{\factind}}}\|\dualadmmmat_{
\pfactmat\factnot{\factind}}\admmiternot{\admmiter}-
\dualadmmmat_{\pfactmat\factnot{\factind}}\admmiternot{\admmiter+1}\|_F^2\\
\lagragian(\dualadmmvec_{\pdiagvec\factnot{\factind}}\admmiternot{\admmiter})
-\lagragian(\dualadmmvec_{\pdiagvec\factnot{\factind}}\admmiternot{\admmiter+1})
&=-\frac{1}{\regadmm_{\pdiagvec\factnot{\factind}}}\|\dualadmmvec_{
\pdiagvec\factnot{\factind}}\admmiternot{\admmiter}-
\dualadmmvec_{\pdiagvec\factnot{\factind}}\admmiternot{\admmiter+1}\|_F^2.
\end{align}
\end{subequations}
Hence, we find that
\begin{align}
\lagragian(\collectedvar\admmiternot{\admmiter+1},\collecteddualvar\admmiternot{\admmiter})-
\lagragian(\collectedvar\admmiternot{\admmiter+1},\collecteddualvar\admmiternot{\admmiter+1})\ge
-\frac{1}{R}\|\collecteddualvar\admmiternot{\admmiter}-\collecteddualvar\admmiternot{\admmiter+1}\|_F^2
\label{eq:dualbound}
\end{align}
and upon combining \eqref{eq:dualbound} and \eqref{eq:primalbound}, we arrive at
\begin{align}
\lagragian(\collectedvar\admmiternot{\admmiter},\collecteddualvar\admmiternot{\admmiter})-
\lagragian(\collectedvar\admmiternot{\admmiter+1},\collecteddualvar\admmiternot{\admmiter+1})\ge&\\
R\|\collectedvar\admmiternot{\admmiter}-\collectedvar\admmiternot{\admmiter+1}\|_F^2&
-\frac{1}{R}\|\collecteddualvar\admmiternot{\admmiter}-\collecteddualvar\admmiternot{\admmiter+1}\|_F^2.\nonumber
\end{align}
Since $\lagragian(\cdot)$ is bounded, we have
\begin{align}
    \sum_{\admmiter=0}^{\infty} 
 R\|\collectedvar\admmiternot{\admmiter}-\collectedvar\admmiternot{\admmiter+1}\|_F^2
-\frac{1}{R}\|\collecteddualvar\admmiternot{\admmiter}-\collecteddualvar\admmiternot{\admmiter+1}\|_F^2
<\infty
\end{align}
and after applying \eqref{eq:thassump} we establish that  $\collectedvar\admmiternot{\admmiter+1}-
\collectedvar\admmiternot\admmiter\rightarrow\mathbf{0}$ and 
$\collecteddualvar\admmiternot{\admmiter+1}-
\collecteddualvar\admmiternot\admmiter\rightarrow\mathbf{0}$. 

Next, we rewrite the ADMM updates in \eqref{eq:admmupd} as
\begin{subequations}
\label{eq:updre}
\begin{align}
&[\factmat\factnot{\factind}\admmiternot{\admmiter+1}-\factmat\factnot{\factind}\admmiternot{\admmiter})
(\katriraofac\factnot{\factind}^\transpose\katriraofac\factnot{\factind}+\regpar 
\diagmat\factnot{\factind}\admmiternot{\admmiter}{\cfactmat\factnot{\factind}\admmiternot{\admmiter}}^\transpose 
\cfactmat\factnot{\factind}\admmiternot{\admmiter}\diagmat\factnot{\factind}\admmiternot{\admmiter}
+(\regadmm_{\pfactmat\factnot{\factind}}+\regadmm_{\cfactmat\factnot{\factind}})\eye] \nonumber
\\=&(\datamat\factnot{\factind}-
\factmat\factnot{\factind}\admmiternot{\admmiter}\katriraofac\factnot{\factind}^\transpose)
\katriraofac\factnot{\factind}+\mu(\graphmat\factnot{\factind}-
\factmat\factnot{\factind}\admmiternot{\admmiter} \diagmat\factnot{\factind}\admmiternot{\admmiter}
{\cfactmat\factnot{\factind}\admmiternot{\admmiter}}^\transpose)
\cfactmat\factnot{\factind}\admmiternot{\admmiter}\diagmat\factnot{\factind}\admmiternot{\admmiter}\nonumber\\
+&\regadmm_{\pfactmat\factnot{\factind}}(\factmat\factnot{\factind}\admmiternot{\admmiter}-
\pfactmat\factnot{\factind}\admmiternot{\admmiter})
+\regadmm_{\cfactmat\factnot{\factind}}(\factmat\factnot{\factind}\admmiternot{\admmiter}
-\cfactmat\factnot{\factind}\admmiternot{\admmiter})
-
 \!\dualadmmmat_{\pfactmat\factnot{\factind}}\admmiternot{\admmiter} - \!\dualadmmmat_{\cfactmat\factnot{\factind}}\admmiternot{\admmiter}\\
 &(\diagvec\factnot{\factind}\admmiternot{\admmiter+1}-\diagvec\factnot{\factind}\admmiternot{\admmiter})
 ((\cfactmat\factnot{\factind}\admmiternot{\admmiter}\odot\factmat\factnot{\factind}\admmiternot{\admmiter})^\transpose
({\regpar}\cfactmat\factnot{\factind}\admmiternot{\admmiter}\odot\factmat\factnot{\factind}\admmiternot{\admmiter})+
\regadmm_{\pdiagvec\factnot{\factind}}\eye)\nonumber\\
=&
{\regpar}(\cfactmat\factnot{\factind}\admmiternot{\admmiter}\odot
\factmat\factnot{\factind}\admmiternot{\admmiter})^\transpose (\graphvec\factnot{\factind}
-	\cfactmat\factnot{\factind}\admmiternot{\admmiter}\odot\factmat\factnot{\factind}\admmiternot{\admmiter}
\pdiagvec\factnot{\factind}\admmiternot{\admmiter})+
\regadmm_{\dualadmmvec_{\pdiagvec\factnot{\factind}}}(\diagvec\factnot{\factind}\admmiternot{\admmiter}
-\pdiagvec\factnot{\factind}\admmiternot{\admmiter})-	\dualadmmvec_{\pdiagvec\factnot{\factind}}
\admmiternot{\admmiter}\\
&(\cfactmat\factnot{\factind}\admmiternot{\admmiter+1} -\cfactmat\factnot{\factind}\admmiternot{\admmiter})(\regpar 
\diagmat\factnot{\factind}\admmiternot{\admmiter}{\factmat\factnot{\factind}\admmiternot{\admmiter}}^\transpose 
\factmat\factnot{\factind}\admmiternot{\admmiter}\diagmat\factnot{\factind}\admmiternot{\admmiter} +\regadmm_{\cfactmat\factnot{\factind}}\eye) \nonumber\\
=&\regpar (\graphmat\factnot{\factind}-
	\cfactmat\factnot{\factind}\admmiternot{\admmiter}\diagmat\factnot{\factind}\admmiternot{\admmiter}
	{\factmat\factnot{\factind}\admmiternot{\admmiter}}^\transpose) 
	\factmat\factnot{\factind}\admmiternot{\admmiter}\diagmat\factnot{\factind}\admmiternot{\admmiter}
+	\regadmm_{\cfactmat\factnot{\factind}}(\factmat\factnot{\factind}\admmiternot{\admmiter}
-\cfactmat\factnot{\factind}\admmiternot{\admmiter})
- \!\dualadmmmat_{\cfactmat\factnot{\factind}}\admmiternot{\admmiter}\\
&\pfactmat\factnot{\factind}\admmiternot{\admmiter+1} -\pfactmat\factnot{\factind}\admmiternot{\admmiter}
=\bigg(\factmat\factnot{\factind}\admmiternot{\admmiter}+\frac{1}{ 
\regadmm_{\pfactmat\factnot{\factind}}}\dualadmmmat_{\pfactmat\factnot{\factind}}\admmiternot{\admmiter}\bigg)_+ 
-\pfactmat\factnot{\factind}\admmiternot{\admmiter}\\
&\pdiagvec\factnot{\factind}\admmiternot{\admmiter+1} -\pdiagvec\factnot{\factind}\admmiternot{\admmiter}
=\bigg(\diagvec\factnot{\factind}\admmiternot{\admmiter}+\frac{1}{ 
\regadmm_{\pdiagvec\factnot{\factind}}}\dualadmmvec_{\pdiagvec\factnot{\factind}}\admmiternot{\admmiter}\bigg)_+ 
-\pdiagvec\factnot{\factind}\admmiternot{\admmiter}
\end{align}
and for the dual updates
\begin{align}
\dualadmmmat_{\cfactmat\factnot{\factind}}\admmiternot{\admmiter+1}-
\dualadmmmat_{\cfactmat\factnot{\factind}}\admmiternot{\admmiter}=&
\regadmm_{\cfactmat\factnot{\factind}}(\factmat\factnot{\factind}\admmiternot{\admmiter}
-\cfactmat\factnot{\factind}\admmiternot{\admmiter})\nonumber\\
\dualadmmmat_{\pfactmat\factnot{\factind}}\admmiternot{\admmiter+1}-\dualadmmmat_{\pfactmat\factnot{\factind}}
\admmiternot{\admmiter}=&
\regadmm_{\pfactmat\factnot{\factind}}(\factmat\factnot{\factind}\admmiternot{\admmiter}-
\pfactmat\factnot{\factind}\admmiternot{\admmiter})\nonumber\\
\dualadmmvec_{\pdiagvec\factnot{\factind}}\admmiternot{\admmiter+1}-
\dualadmmvec_{\pdiagvec\factnot{\factind}}\admmiternot{\admmiter}=& 
\regadmm_{\pdiagvec\factnot{\factind}}(\diagvec\factnot{\factind}\admmiternot{\admmiter}
-\pdiagvec\factnot{\factind}\admmiternot{\admmiter}).
\label{eq:lagrangupdre}
\end{align}
\end{subequations}
Next, we leverage \eqref{eq:conv} and establish that the left hand side of
the equations in \eqref{eq:updre} is equal to $\mathbf{0}$. Hence, from \eqref{eq:lagrangupdre}
we deduce that $\factmat\factnot{\factind}\admmiternot{\admmiter}
-\cfactmat\factnot{\factind}\admmiternot{\admmiter}\rightarrow\mathbf{0}$, $\factmat\factnot{\factind}\admmiternot{\admmiter}
-\cfactmat\factnot{\factind}\admmiternot{\admmiter}\rightarrow\mathbf{0}$, and 
$\factmat\factnot{\factind}\admmiternot{\admmiter}
-\cfactmat\factnot{\factind}\admmiternot{\admmiter}\rightarrow\mathbf{0}$. 
So far we have proved that the 
KKT conditions \eqref{eq:kkt} relating to the primal variables $\collectedvar$,
are satisfied.  The variables $\pfactmat\factnot{\factind}$ and 
$\pdiagvec\factnot{\factind}$ are nonnegative by construction. For the dual variables, notice from \eqref{eq:lagrangupdre}
that if $[\factmat\factnot{\factind}\admmiternot{\admmiter}]_{(\tensind\factnot{\factind},\ranktensorind)}=
[\pfactmat\factnot{\factind}\admmiternot{\admmiter}]_{(\tensind\factnot{\factind},\ranktensorind)}=0$ then 
$\big([\dualadmmmat_{\pfactmat\factnot{\factind}}\admmiternot{\admmiter}]_{(\tensind\factnot{\factind},\ranktensorind)}\big)_+=0$, which implies that $[\dualadmmmat_{\pfactmat\factnot{\factind}}\admmiternot{\admmiter}]_
{(\tensind\factnot{\factind},\ranktensorind)}\le{0}$, else if $[\factmat\factnot{\factind}\admmiternot{\admmiter}]_{(\tensind\factnot{\factind},\ranktensorind)}=
[\pfactmat\factnot{\factind}\admmiternot{\admmiter}]_{(\tensind\factnot{\factind},\ranktensorind)}\ge0$
then $[\dualadmmmat_{\pfactmat\factnot{\factind}}\admmiternot{\admmiter}]_
{(\tensind\factnot{\factind},\ranktensorind)}=0$. The same argument applies for 
$\dualadmmvec_{\pdiagvec\factnot{\factind}}\admmiternot{\admmiter}$ and hence we have established satisfaction of 
the last KKT conditions concerning $\dualadmmmat_{\pfactmat\factnot{\factind}}\admmiternot{\admmiter}$
and $\dualadmmvec_{\pdiagvec\factnot{\factind}}\admmiternot{\admmiter}$.

\bibliographystyle{IEEEtran}
\bibliography{IEEEabrv,my_bibliography}
\noindent

\begin{IEEEbiography}[{\includegraphics[width=1in,height=1.25in,clip,keepaspectratio]{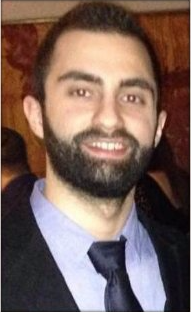}}]{Vassilis
 N. Ioannidis (S'16)} 
 received his Diploma in Electrical and Computer
Engineering from the National Technical University of Athens, Greece,
in 2015, and the M.Sc. degree in Electrical Engineering
from the University of Minnesota (UMN),
Twin Cities, Minneapolis, MN, USA, in 2017. Currently, he is working towards 
the Ph.D. degree in the
Department of Electrical and Computer Engineering at the
University of Minnesota, Twin Cities. Vassilis received the Doctoral Dissertation Fellowship (2019) from the University of Minnesota.  He also received Student Travel Awards from the IEEE Signal
Processing Society (2017,2018) and from the IEEE (2018). From 2014 to 2015, he worked as
a middleware consultant for Oracle in Athens, Greece, and received a
Performance Excellence award. His research interests include
machine learning, big data analytics, and network science.
\end{IEEEbiography}
 
\begin{IEEEbiography}[{\includegraphics[width=1in,height=1.25in,clip,keepaspectratio]{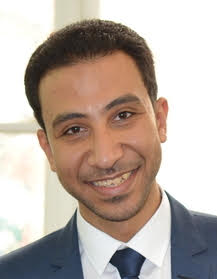}}]{
Ahmed S. Zamzam (S'14)} is a PhD Candidate at the Department of Electrical and Computer Engineering at the University of Minnesota, where he is also affiliated with the Signal and Tensor Analytics Research (STAR) group under the supervision of Professor N. D. Sidiropoulos. Previously, he earned his BSc at Cairo University in 2013. In 2015, he received the MSc from Nile University. 	
Ahmed received the Louis John Schnell Fellowship (2015), and the Doctoral Dissertation Fellowship (2018) from the University of Minnesota. He also received Student Travel Awards from the IEEE Signal Processing Society (2017), the IEEE Power and Energy Society (2018), and the Council of Graduate Students at the University of Minnesota (2016, 2018). His research interests include control and optimization of smart grids, large-scale complex energy systems, grid data analytics, and machine learning.  \end{IEEEbiography}

\begin{IEEEbiography}[{\includegraphics[width=1in,height=1.25in,clip,keepaspectratio]{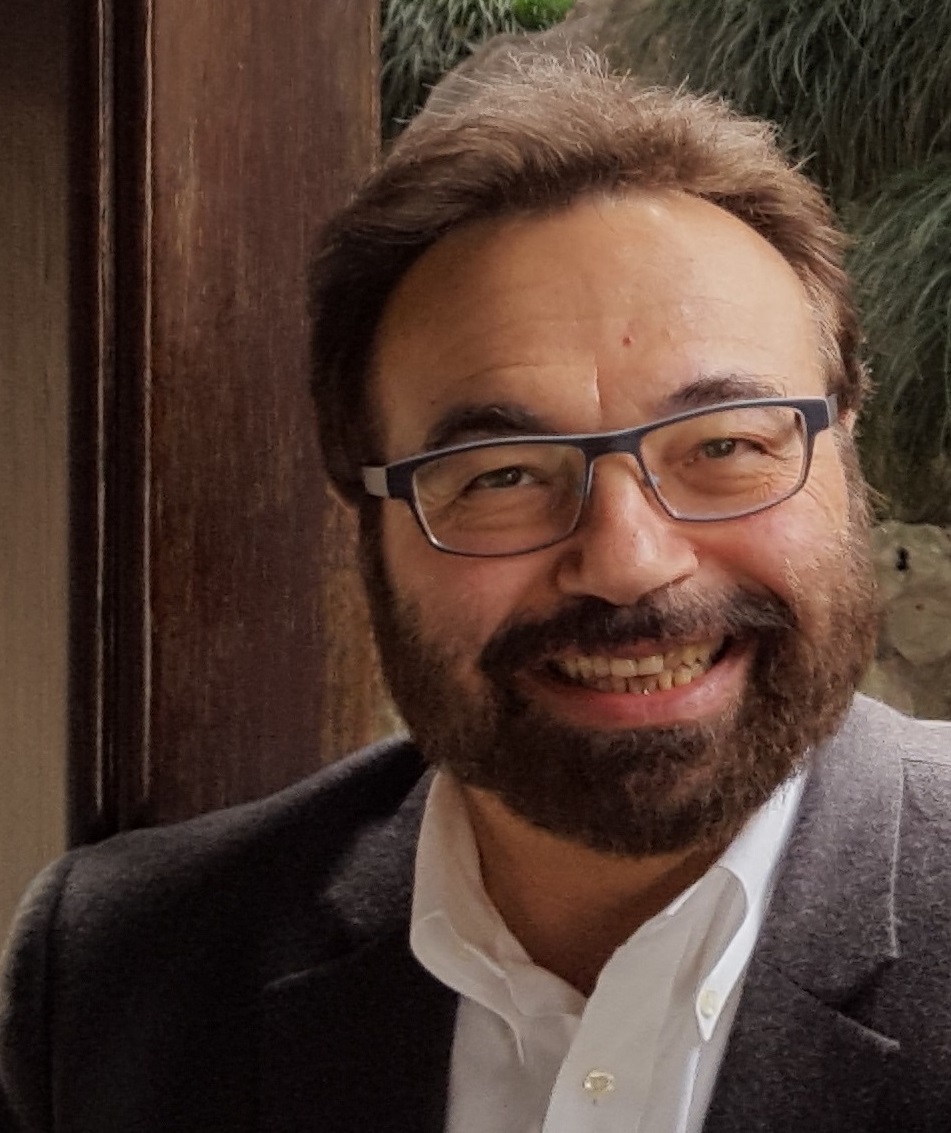}}]{G. B.  Giannakis} (Fellow'97) received his Diploma in Electrical
Engr. from the Ntl. Tech. Univ. of Athens, Greece, 1981. From
1982 to 1986 he was with the Univ. of Southern California (USC),
where he received his MSc. in Electrical Engineering, 1983, MSc.
in Mathematics, 1986, and Ph.D. in Electrical Engr., 1986. He
was a faculty member with the University of Virginia from 1987
to 1998, and since 1999 he has been a professor with the Univ.
of Minnesota, where he holds an ADC Endowed Chair, a University
of Minnesota McKnight Presidential Chair in ECE, and serves as
director of the Digital Technology Center.

His general interests span the areas of statistical learning,
communications, and networking - subjects on which he has published
more than 450 journal papers, 750 conference papers, 25 book
chapters, two edited books and two research monographs (h-index 140).
Current research focuses on Data Science, and Network Science with
applications to the Internet of Things, social, brain, and power
networks with renewables. He is the (co-) inventor of 32 patents issued,
and the (co-) recipient of 9 best journal paper awards from the
IEEE Signal Processing (SP) and Communications Societies, including
the G. Marconi Prize Paper Award in Wireless Communications. He also
received Technical Achievement Awards from the SP Society (2000),
from EURASIP (2005), a Young Faculty Teaching Award, the G. W. Taylor
Award for Distinguished Research from the University of Minnesota,
and the IEEE Fourier Technical Field Award (inaugural recipient in
2015). He is a Fellow of EURASIP, and has served the IEEE in a number
of posts, including that of a Distinguished Lecturer for the IEEE-SPS.
\end{IEEEbiography}

\vfill
\begin{IEEEbiography}[{\includegraphics[width=1in,height=1.25in,clip,keepaspectratio]{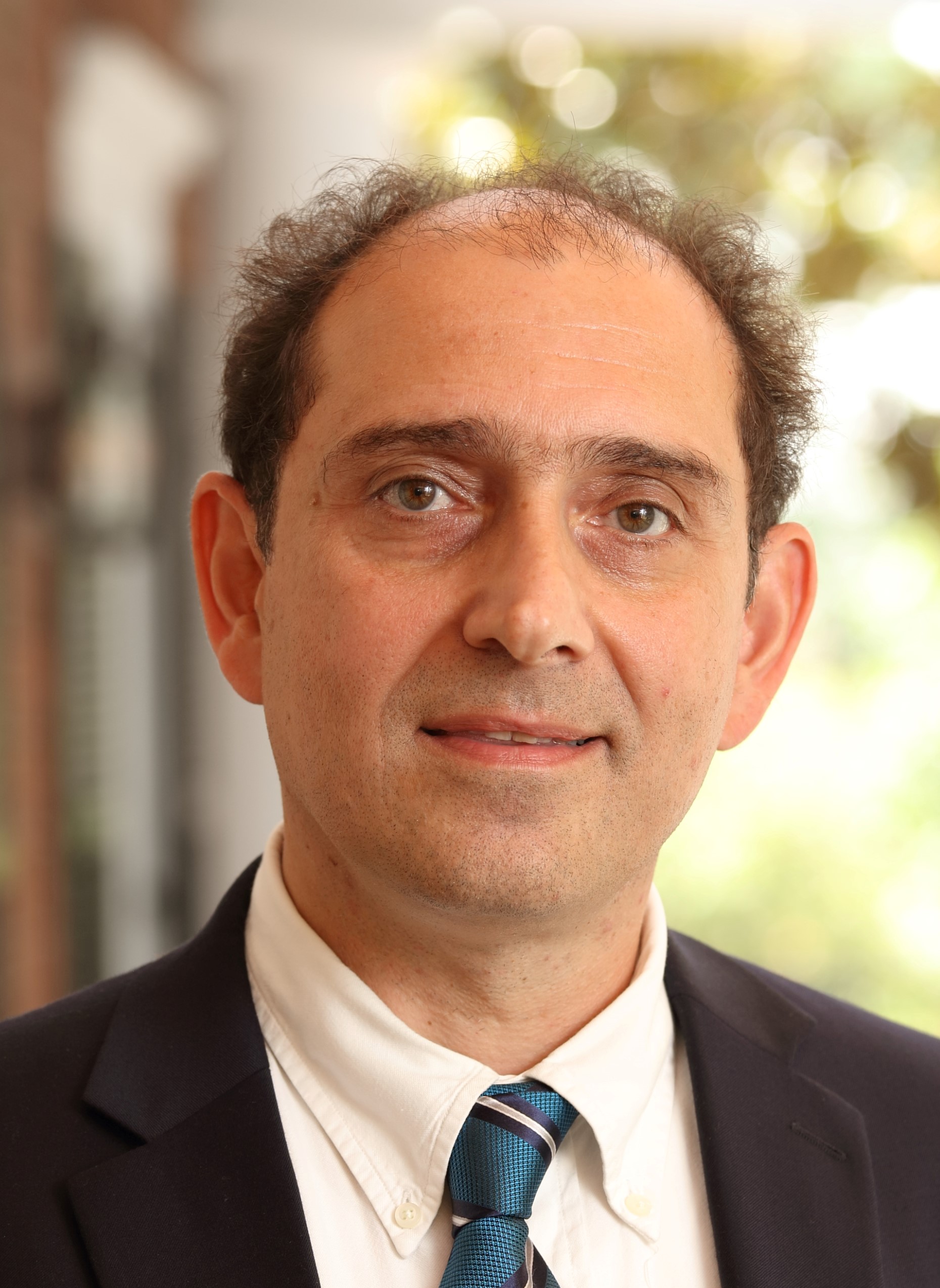}}]{
Nicholas D. Sidiropoulos (F'09)}earned the Diploma in Electrical Engineering from Aristotle University of Thessaloniki, Greece, and M.S. and Ph.D. degrees in Electrical Engineering from the University of Maryland at College Park, in 1988, 1990 and 1992, respectively. He has served on the faculty of the University of Virginia, University of Minnesota, and the Technical University of Crete, Greece, prior to his current appointment as Louis T. Rader Professor and Chair of ECE at UVA. From 2015 to 2017 he was an ADC Chair Professor at the University of Minnesota. His research interests are in signal processing, communications, optimization, tensor decomposition, and factor analysis, with applications in machine learning and communications. He received the NSF/CAREER award in 1998, the IEEE Signal Processing Society (SPS) Best Paper Award in 2001, 2007, and 2011, served as IEEE SPS Distinguished Lecturer (2008-2009), and currently serves as Vice President - Membership of IEEE SPS. He received the 2010 IEEE Signal Processing Society Meritorious Service Award, and the 2013 Distinguished Alumni Award from the University of Maryland, Dept. of ECE. He is a Fellow of IEEE (2009) and a Fellow of EURASIP (2014).   \end{IEEEbiography}

\end{document}